\documentclass[runningheads]{llncs}
\usepackage{graphicx}
\usepackage{amsmath,amssymb} % define this before the line numbering.
\usepackage{color}
\usepackage{array}
\usepackage{multirow}
\usepackage{subfigure}
\usepackage{makecell}

\usepackage[width=122mm,left=12mm,paperwidth=146mm,height=193mm,top=12mm,paperheight=217mm]{geometry}
\begin{document}
% \linenumbers
\pagestyle{headings}
\mainmatter
\def\ECCV18SubNumber{}  % Insert your submission number here

\title{Depth-aware CNN for RGB-D Segmentation} % Replace with your title

\titlerunning{Depth-aware CNN for RGB-D Segmentation}

\authorrunning{Wang and Neumann}

\author{Weiyue Wang\quad Ulrich Neumann}
\institute{University of Southern California\\
Los Angeles, California}

\maketitle

\begin{abstract}
Convolutional neural networks (CNN) are limited by the lack of capability to handle geometric information due to the fixed grid kernel structure. The availability of depth data enables progress in RGB-D semantic segmentation with CNNs. State-of-the-art methods either use depth as additional images or process spatial information in 3D volumes or point clouds. These methods suffer from high computation and memory cost. To address these issues, we present Depth-aware CNN by introducing two intuitive, flexible and effective operations: depth-aware convolution and depth-aware average pooling. By leveraging depth similarity between pixels in the process of information propagation, geometry is seamlessly incorporated into CNN. Without introducing any additional parameters, both operators can be easily integrated into existing CNNs. Extensive experiments and ablation studies on challenging RGB-D semantic segmentation benchmarks validate the effectiveness and flexibility of our approach. 
\end{abstract}

\section{Introduction}
Recent advances~\cite{fcn,dilated,deeplab} in CNN have achieved significant success in scene understanding. With the help of range sensors (such as Kinect, LiDAR etc.), depth images are applicable along with RGB images. Taking advantages of the two complementary modalities with CNN is able to improve the performance of scene understanding. However, CNN is limited to model geometric variance due to the fixed grid computation structure. Incorporating the geometric information from depth images into CNN is important yet challenging. 

Extensive studies~\cite{xiaojuaniccv17,localitysensitive,eccvkhan,Silberman:ECCV12,conf/cvpr/RenBF12,couprie2013indoor,wang2017selfpaced} have been carried out on this task. FCN~\cite{fcn} and its successors treat depth as another input image and construct two CNNs to process RGB and depth separately. This doubles the number of network parameters and computation cost. In addition, the two-stream network architecture still suffers from the fixed geometric structures of CNN. Even if the geometric relations of two pixels are given, this relation cannot be used in information propagation of CNN. 
An alternative is to leverage 3D networks~\cite{xiaojuaniccv17,ssc,wang2017shapeinpainting} to handle geometry. Nevertheless, both volumetric CNNs~\cite{ssc} and 3D point cloud graph networks~\cite{xiaojuaniccv17} are computationally more expensive than 2D CNN. Despite the encouraging results of these progresses, we need to seek a more flexible and efficient way to exploit 3D geometric information in 2D CNN.

\begin{figure}[tb]
	\centering
	\includegraphics[width=0.8\textwidth]{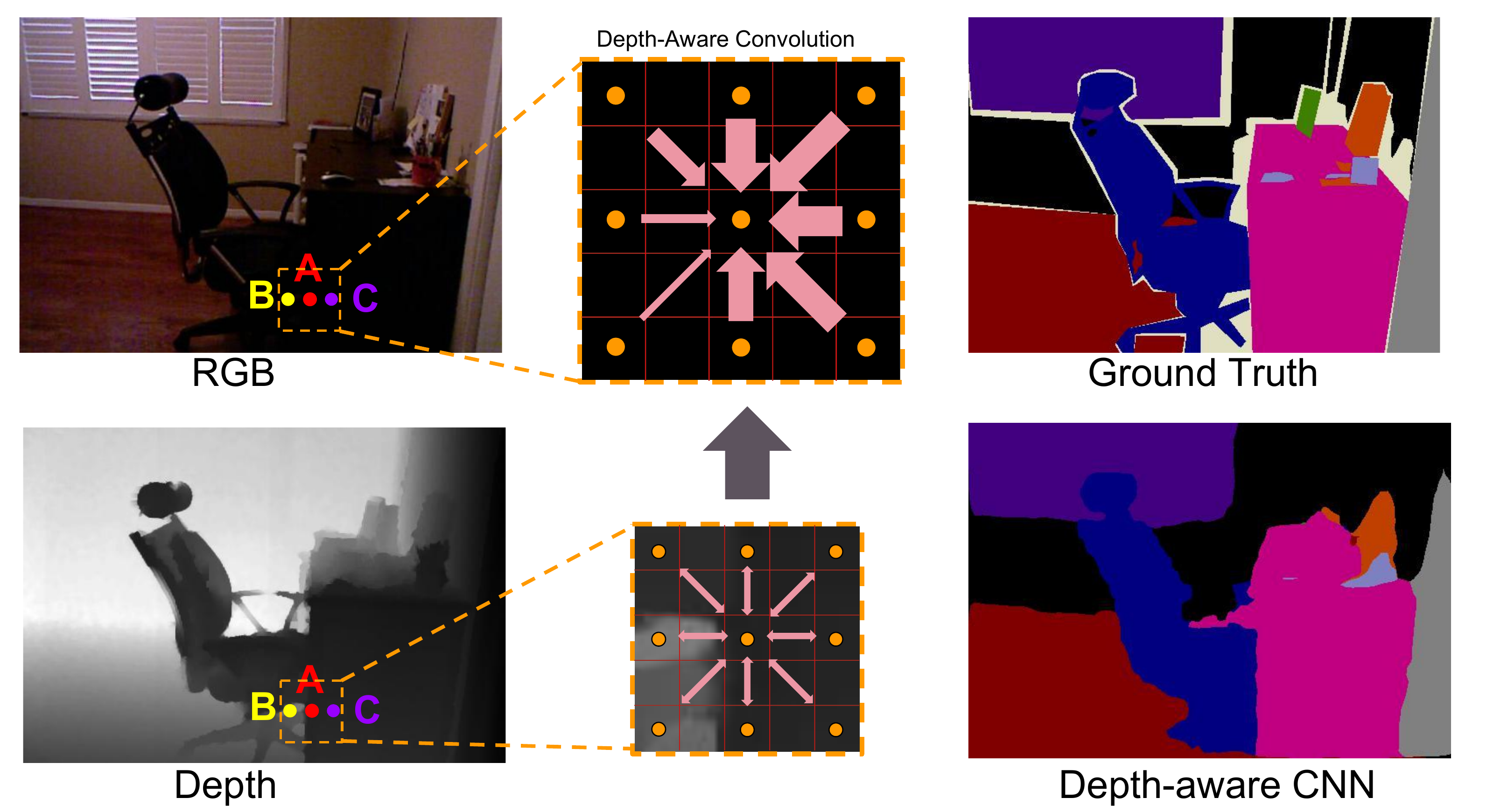}\vspace{-5pt}
	\caption{Illustration of Depth-aware CNN. A and C are labeled as table and B is labeled as chair. They all have similar visual features in the RGB image, while they are separable in depth. Depth-aware CNN incorporate the geometric relations of pixels in both convolution and pooling. When A is the center of the receptive field, C then has more contribution to the output unit than B. Figures in the rightmost column shows the RGB-D semantic segmentation result of Depth-aware CNN. }\vspace{-12pt}
	\label{fig:teaser}
\end{figure}

To address the aforementioned problems, in this paper, we present an end-to-end network, Depth-aware CNN (D-CNN), for RGB-D segmentation. Two new operators are introduced: \textit{depth-aware convolution} and \textit{depth-aware average pooling}. \textit{Depth-aware convolution} augments the standard convolution with a depth similarity term. We force pixels with similar depths with the center of the kernel to have more contribution to the output than others. This simple depth similarity term efficiently incorporates geometry in a convolution kernel and helps build a depth-aware receptive field, where convolution is not constrained to the fixed grid geometric structure.

The second introduced operator is \textit{depth-ware average pooling}. Similarly, when a filter is applied on a local region of the feature map, the pairwise relations in depth between neighboring pixels are considered in computing mean of the local region. Visual features are able to propagate along with the geometric structure given in depth images. Such geometry-aware operation enables the localization of object boundaries with depth images.

Both operators are based on the intuition that pixels with the same semantic label and similar depths should have more impact on each other. We observe that two pixels with the same semantic labels have similar depths. As illustrated in Figure~\ref{fig:teaser}, pixel A and pixel C should be more correlated with each other than pixel A and pixel B. This correlation difference is obvious in depth image while it is not captured in RGB image. By encoding the depth correlation in CNN, pixel C has more contribution to the output unit than pixel B in the process of information propagation.
 
The main advantages of depth-aware CNN are summarized as follows:
\begin{itemize}
	\item By exploiting the nature of CNN kernel handling spatial information, geometry in depth image is able to be integrated into CNN seamlessly.
	\item Depth-aware CNN does not introduce any parameters and computation complexity to the conventional CNN.
	\item Both \textit{depth-aware convolution} and \textit{depth-ware average pooling} can replace their standard counterparts in conventional CNNs with minimal cost.
\end{itemize}

Depth-aware CNN is a general framework that bonds 2D CNN and 3D geometry. Comparison with the state-of-the-art methods and extensive ablation studies on RGB-D semantic segmentation illustrate the flexibility, efficiency and effectiveness of our approach.

%------------------------------------------------------------------------
\section{Related Works}
\subsection{RGB-D Semantic Segmentation}
With the help of CNNs, semantic segmentation on 2D images have achieved promising results~\cite{fcn,dilated,deeplab,huang2016scenelabeling}. These advances in 2D CNN and the availability of depth sensors enables progresses in RGB-D segmentation. Compared to the RGB settings, RGB-D segmentation is able to integrate geometry into scene understanding. In \cite{eigeniccv15,lingni17iros,fusenet2016accv,wang2016eccv}, depth is simply treated as additional channels and directly fed into CNN. Several works~\cite{fcn,fusenet2016accv,eccv14hha,lstmcf,Park_2017_ICCV} encode depth to HHA image, which has three channels: horizontal disparity, height above ground, and norm angle. RGB image and HHA image are fed into two separate networks, and the two predictions are summed up in the last layer. The two-stream network doubles the number of parameters and forward time compared to the conventional 2D network. Moreover, CNNs per se are limited in their ability to model geometric transformations due to their fixed grid computation. Cheng et al.~\cite{localitysensitive} propose a locality-sensitive deconvolution network with gated fusion. They build a feature affinity matrix to perform weighted average pooling and unpooling. Lin et al.~\cite{Lin_2017_ICCV} discretize depth and build different branches for different discrete depth value. He et al.~\cite{yang_cvpr17} use spatio-temporal correspondences across frames to aggregate information over space and time. This requires heavy pre and post-processing such as optical flow and superpixel computation.

Alternatively, many works~\cite{ssc,DeepSlidingShapes} attempt to solve the problem with 3D CNNs. However, the volumetric representation prevents scaling up due to high memory and computation cost. Recently, deep learning frameworks~\cite{xiaojuaniccv17,pointnet,pointnet++,wang2018sgpn,huang2018rsnet} on point cloud are introduced to address the limitations of 3D volume. Qi et al.~\cite{xiaojuaniccv17} built a 3D k-nearest neighbor (kNN) graph neural network on a point cloud with extracted features from a CNN and achieved the state-of-the-art on RGB-D segmentation. Although their method is more efficient than 3D CNNs, the kNN operation suffers from high computation complexity and lack of flexibility. Instead of using 3D representations, we use the raw depth input and integrate 3D geometry into 2D CNN in a more efficient and flexible fashion. 

\subsection{Spatial Transformations in CNN}

Standard CNNs are limited to model geometric transformations due to the fixed structure of convolution kernels. Recently, many works are focused on dealing with this issue. Dilated convolutions~\cite{dilated,deeplab} increases the receptive field size with keeping the same complexity in parameters. This operator achieves better performance on vision tasks such as semantic segmentation. Spatial transform networks~\cite{stn} warps feature maps with a learned global spatial transformation. Deformable CNN~\cite{deformable} learns kernel offsets to augment the spatial sampling locations. These methods have shown geometric transformations enable performance boost on different vision tasks.

With the advances in 3D sensors, depth is applicable at low cost. The geometric information that resides in depth is highly correlated with the spatial transformation in CNN. Our method integrates the geometric relation of pixels into basic operations of CNN, i.e. convolution and pooling, where we use a weighted kernel and force every neuron to have different contributions to the output. This weighted kernel is defined by depth and is able to incorporate geometric relationships without introducing any parameter.
\section{Depth-aware CNN}
In this section, we introduce two depth-aware operations: depth-aware convolution and depth-aware average pooling. They are both simple and intuitive. Both operations require two inputs: the input feature map $\mathbf{x} \in \mathbb{R}^{c_i\times h \times w}$ and the depth image $\mathbf{D}\in \mathbb{R}^{h \times w}$, where $c_i$ is the number of input feature channels, $h$ is the height and $w$ is the width. The output feature map is denoted as $\mathbf{y} \in \mathbb{R}^{c_o\times h \times w}$, where $c_o$ is the number of output feature channels. Although $\mathbf{x}$ and $\mathbf{y}$ are both 3D tensors, the operations are explained in 2D spatial domain for notation clarity and they remain the same across different channels. 

\begin{figure}
	\centering
	\begin{tabular}{cc}
		\includegraphics[width=0.4\textwidth]{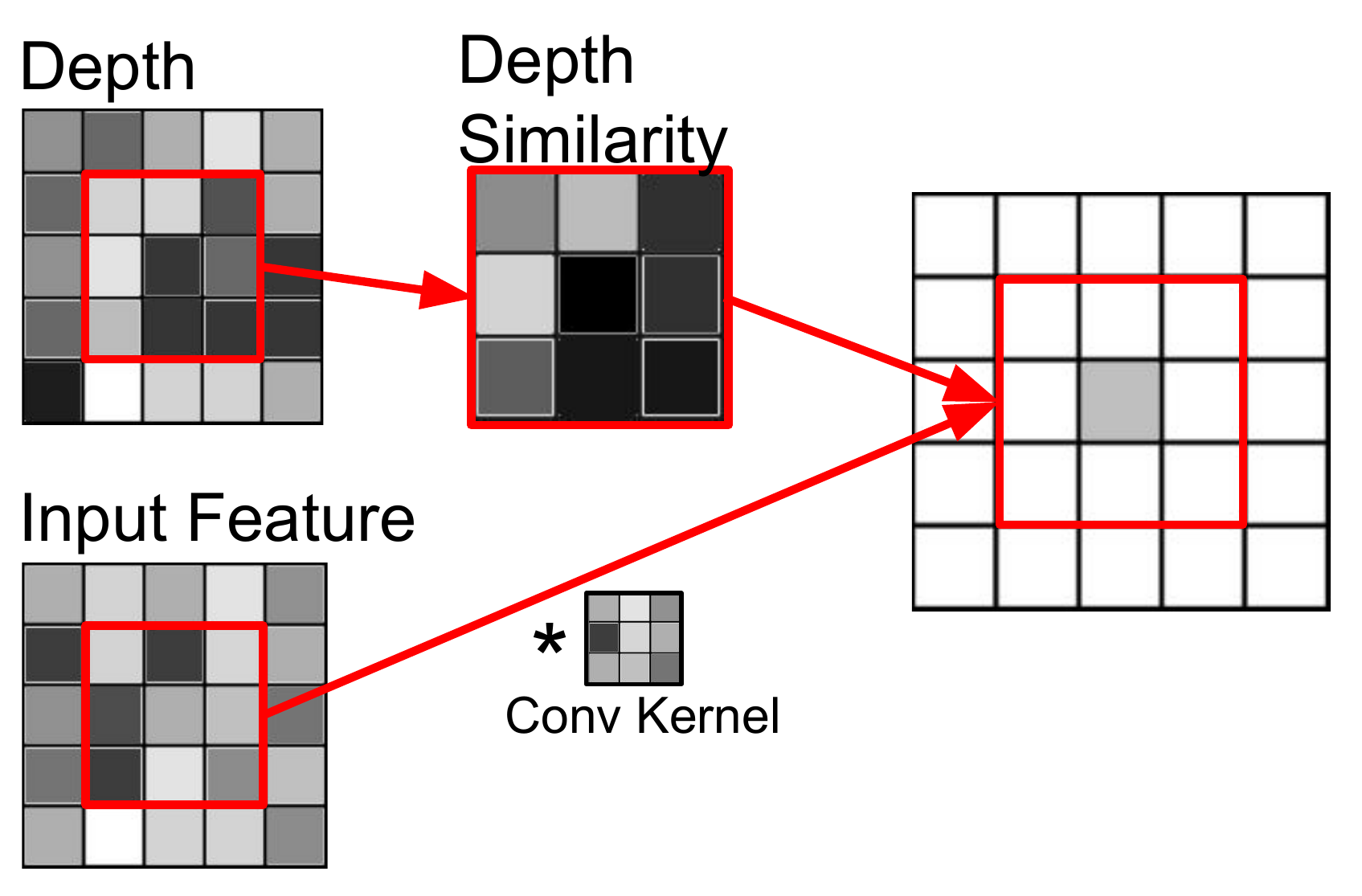}
		&
		\includegraphics[width=0.4\textwidth]{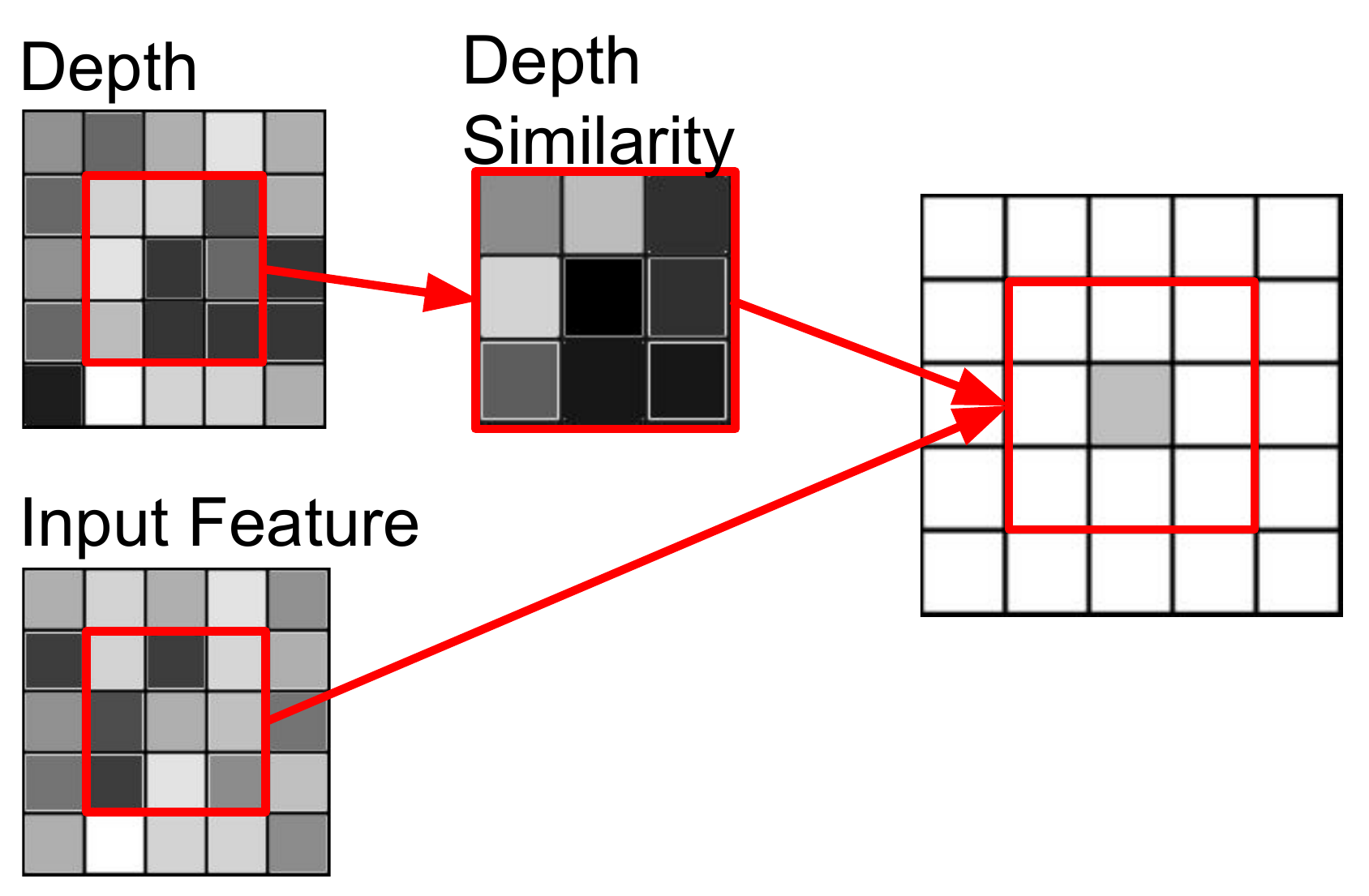}\\
		(a) Depth-aware Convolution &(b) Depth-aware Average Pooling
	\end{tabular}
	\caption{Illustration of information propagation in Depth-aware CNN. Without loss of generality, we only show one filter window with kernel size $3\times 3$. In depth similarity shown in figure, darker color indicates higher similarity while lighter color represents that two pixels are less similar in depth. In (a), the output activation of depth-aware convolution is the multiplication of depth similarity window and the convolved window on input feature map. Similarly in (b), the output of depth-aware average pooling is the average value of the input window weighted by the depth similarity. }\vspace{-10pt}
	\label{fig:depthconv}
\end{figure}
\subsection{Depth-aware Convolution}
A standard 2D convolution operation is the weighted sum of a local grid. For each pixel location $\boldsymbol{p_0}$ on $\mathbf{y}$, the output of standard 2D convolution is

\begin{equation}
\mathbf{y}(\mathbf{p}_0)=\sum_{\mathbf{p}_n\in\mathcal{R}}\mathbf{w}(\mathbf{p}_n)\cdot \mathbf{x}(\mathbf{p}_0+\mathbf{p}_n),
\label{eq:standard_conv}
\end{equation}
where $\mathcal{R}$ is the local grid around $\boldsymbol{p_0}$ in $\mathbf{x}$ and $\mathbf{w}$ is the convolution kernel. $\mathcal{R}$ can be a regular grid defined by kernel size and dilation~\cite{dilated}, and it can also be a non-regular grid~\cite{deformable}.

As is shown in Figure~\ref{fig:teaser}, pixel A and pixel B have different semantic labels and different depths while they are not separable in RGB space. On the other hand, pixel A and pixel C have the same labels and similar depths. To exploit the depth correlation between pixels, depth-aware convolution simply adds a depth similarity term, resulting in two sets of weights in convolution: 1) the learnable convolution kernel $\mathbf{w}$; 2) depth similarity $F_\mathbf{D}$ between two pixels. Consequently, Equ.~\ref{eq:standard_conv} becomes

\begin{equation}
\mathbf{y}(\mathbf{p}_0)=\sum_{\mathbf{p}_n\in\mathcal{R}}\mathbf{w}(\mathbf{p}_n)\cdot
F_\mathbf{D}(\mathbf{p}_0,\mathbf{p}_0+\mathbf{p}_n)\cdot
\mathbf{x}(\mathbf{p}_0+\mathbf{p}_n).
\label{eq:depthconv}
\end{equation}
And $F_\mathbf{D}(\mathbf{p}_i,\mathbf{p}_j)$ is defined as 

\begin{equation}
F_\mathbf{D}(\mathbf{p}_i,\mathbf{p}_j)=
\exp(-\alpha|\mathbf{D}(\mathbf{p}_i) - \mathbf{D}(\mathbf{p}_j)|),
\label{eq:fd}
\end{equation}
where $\alpha$ is a constant. The selection of $F_\mathbf{D}$ is based on the intuition that pixels with similar depths should have more impact on each other. We will study the effect of different $\alpha$ and different $F_\mathbf{D}$ in Section~\ref{sec:ablation}.

The gradients for $\mathbf{x}$ and $\mathbf{w}$ are simply multiplied by $F_\mathbf{D}$. Note that the $F_\mathbf{D}$ part does not require gradient during back-propagation, therefore, Equ.~\ref{eq:depthconv} does not integrate any parameters by the depth similarity term.

Figure~\ref{fig:depthconv}(a) illustrates this process. Pixels which have similar depths with the convolving center will have more impact on the output during convolution.

\subsection{Depth-aware Average Pooling}
The conventional average pooling computes the mean of a grid $\mathcal{R}$ over $\mathbf{x}$. It is defined as

\begin{equation}
\mathbf{y}(\mathbf{p}_0)=\frac{1}{|\mathcal{R}|}
\sum_{\mathbf{p}_n\in\mathcal{R}}\mathbf{x}(\mathbf{p}_0+\mathbf{p}_n).
\label{eq:avgpooling}
\end{equation} 

It treats every pixel equally and will make the object boundary blurry. Geometric information is useful to address this issue. 

Similar to as in depth-aware convolution, we take advantage of the depth similarity $F_\mathbf{D}$ to force pixels with more consistent geometry to make more contribution on the corresponding output. For each pixel location $\boldsymbol{p_0}$, the depth-aware average pooling operation then becomes 

\begin{equation}
\mathbf{y}(\mathbf{p}_0)=\frac{1}{
\sum_{\mathbf{p}_n\in\mathcal{R}}F_\mathbf{D}(\mathbf{p}_0,\mathbf{p}_0+\mathbf{p}_n)}
\sum_{\mathbf{p}_n\in\mathcal{R}} F_\mathbf{D}(\mathbf{p}_0,\mathbf{p}_0+\mathbf{p}_n)\cdot\mathbf{x}(\mathbf{p}_0+\mathbf{p}_n).
\label{eq:depthavgpooling}
\end{equation} 

The gradient should be multiplied by $\frac{F_\mathbf{D}}{\sum_{\mathbf{p}_n\in\mathcal{R}}F_\mathbf{D}(\mathbf{p}_0,\mathbf{p}_0+\mathbf{p}_n)}$ during back propagation. As illustrated in Figure~\ref{fig:depthconv}(b), this operation prevent suffering from the fixed geometric structure of standard pooling.

\subsection{Understanding Depth-aware CNN}
A major advantage of CNN is its capability of using GPU to perform parallel computing and accelerate the computation. This acceleration mainly stems from unrolling convolution operation with the grid computation structure. However, this limits the ability of CNN to model geometric variations. Researchers in 3D deep learning have focused on modeling geometry in deep neural networks in the last few years. As the volumetric representation~\cite{ssc,DeepSlidingShapes} is of high memory and computation cost, point clouds are considered as a more proper representation. However, deep learning frameworks~\cite{pointnet++,xiaojuaniccv17} on point cloud are based on building kNN. This not only suffers from high computation complexity, but also breaks the pixel-wise correspondence between RGB and depth, which makes the framework is not able to leverage the efficiency of CNN's grid computation structure. Instead of operating on 3D data, we exploit the raw depth input. By augmenting the convolution kernel with a depth similarity term, depth-aware CNN captures geometry with transformable receptive field.

Many works have studied spatial transformable receptive field of CNN. Dilated convolution~\cite{deeplab,dilated} has demonstrated that increasing receptive field boost the performance of networks. In deformable CNN~\cite{deformable}, Dai et al. demonstrate that learning receptive field adaptively can help CNN achieve better results. They also show that pixels within the same object in a receptive field contribute more to the output unit than pixels with different labels. We observe that semantic labels and depths have high correlations. Table~\ref{table:var} reports the statistics of pixel depth variance within the same class and across different classes on NYUv2~\cite{nyuv2} dataset. Even the pixel depth variances of large objects such as wall and floor are much smaller than the variance of a whole scene. This indicates that pixels with the same semantic labels tend to have similar depths. This pattern is integrated in Equ.~\ref{eq:depthconv} and Equ.~\ref{eq:depthavgpooling} with $F_\mathbf{D}$. Without introducing any parameter, depth-aware convolution and depth-aware average pooling are able to enhance the localization ability of CNN. We evaluate the impact on performance of different depth similarity functions $F_\mathbf{D}$ in Section~\ref{sec:ablation}.

\begin{table}
	\begin{center}
		\begin{tabular}{c|cccccc}
			\Xhline{3\arrayrulewidth}
			& Wall& Floor & Bed& Chair & Table & All\\
			\hline
			Variance&0.57&0.65&0.12 &0.23&0.34&1.20\\
			\Xhline{3\arrayrulewidth}
		\end{tabular}
	\end{center}
	\caption{Mean depth variance of different categories on NYUv2 dataset. ``All" denotes the mean variance of all categories. For every image, pixel-wise variance of depth for each category is calculated. Averaged variance is then computed over all images. For ``All", all pixels in a image are considered to calculate the depth variance. Mean variance over all images is further computed.}
	\label{table:var}
\end{table}

\begin{figure}[tb]
	\centering
	%\renewcommand{\arraystretch}{0.1}% Tighter
	%\newcolumntype{C}{>{\centering\arraybackslash}p{3.6em}}
	\begin{tabular}{cccc}
		\includegraphics[width=0.23\textwidth]{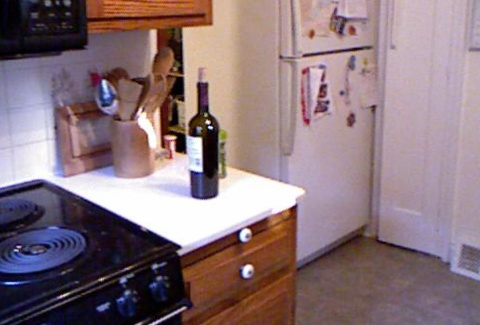}
		&\includegraphics[width=0.23\textwidth]{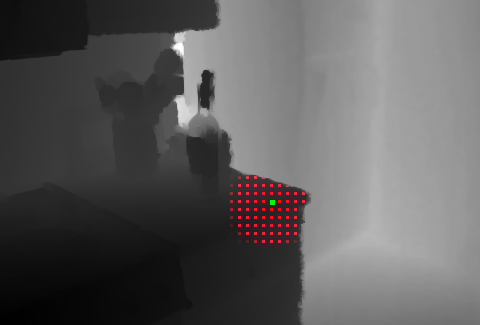}
		&\includegraphics[width=0.23\textwidth]{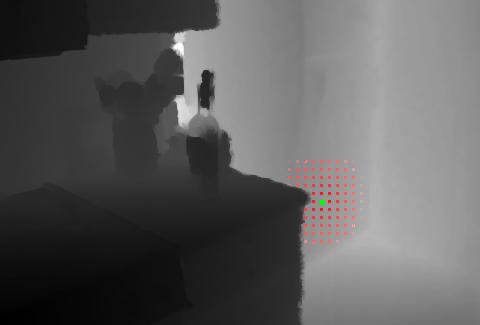}
		&\includegraphics[width=0.23\textwidth]{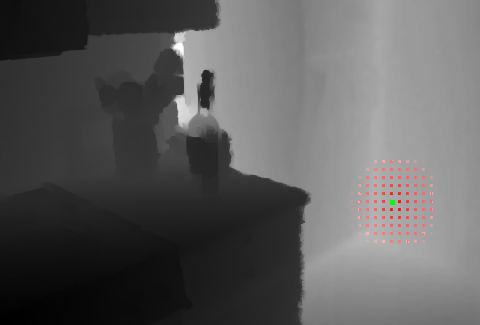}
		\\
		\includegraphics[width=0.23\textwidth]{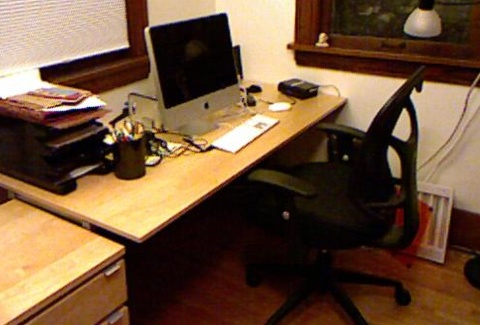}
		&\includegraphics[width=0.23\textwidth]{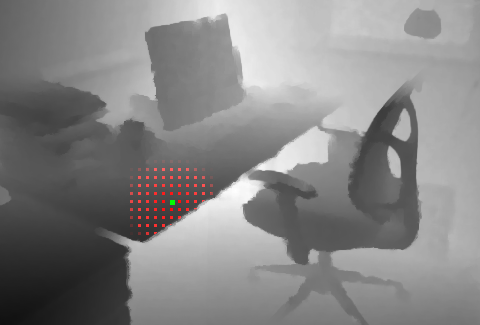}
		&\includegraphics[width=0.23\textwidth]{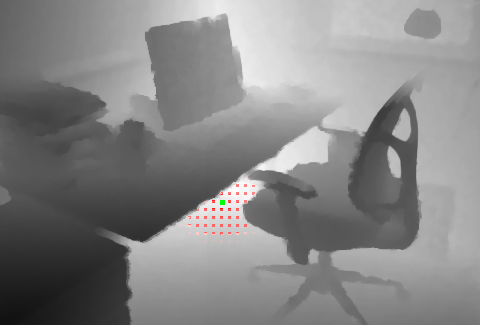}
		&\includegraphics[width=0.23\textwidth]{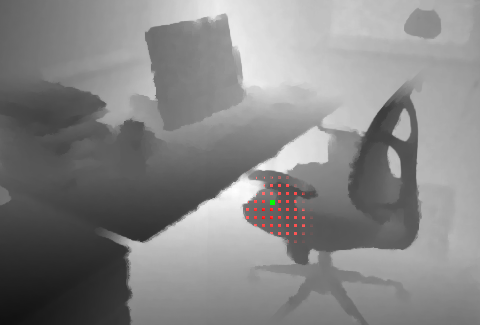}
		\\
		(a) & (b) & (c) & (d)
	\end{tabular}
	\caption{Illustration of effective receptive field of Depth-aware CNN. (a) is the input RGB images. (b), (c) and (d) are depth images. For (b), (c) and (d), we show the sampling locations (red dots) in three levels of $3\times 3$ depth-aware convolutions for the activation unit (green dot).}\vspace{-20pt}
	\label{fig:receptive}
\end{figure}

To get a better understanding of how depth-aware CNN captures geometry with depth, Figure~\ref{fig:receptive} shows the effective receptive field of the given input neuron. In conventional CNN, the receptive fields and sampling locations are fixed across feature map. With the depth-aware term incorporated, they are adjusted by the geometric variance. For example, in the second row of Figure~\ref{fig:receptive}(d), the green point is labeled as chair and the effective receptive field of the green point are essentially chair points. This indicates that the effective receptive field mostly have the same semantic label as the center. This pattern increases CNN's performance on semantic segmentation.

\vspace{-10pt}
\subsection{Depth-aware CNN for RGB-D Semantic Segmentation}
\label{sec:netarch}

In this paper, we focus on RGB-D semantic segmentation with depth-aware CNN. Given an RGB image along with depth, our goal is to produce a semantic mask indicating the label of each pixel. Both depth-aware convolution and average pooling easily replace their counterpart in standard CNN.

\begin{table}
	\begin{center}
		\begin{tabular}{c|c|c|c|c|c|c}
			\Xhline{3\arrayrulewidth}
			layer name & conv1\_x & conv2\_x &conv3\_x &conv4\_x&conv5\_x& conv6 \& conv7\\
			\hline
			\multirow{ 4}{*}{\thead{Baseline \\ DeepLab}}& C3-64-1& C3-128-1 & C3-256-1 & C3-512-1 & C3-512-2 &C3-1024-12\\
			& C3-64-1& C3-128-1 & C3-256-1 & C3-512-1 & C3-512-2 &C1-1024-0\\
			& maxpool   & maxpool    & C3-256-1 & C3-512-1 & C3-512-2  &globalpool+concat\\
			&           &            & maxpool     & maxpool   & avgpool   &\\
			\hline
			\multirow{ 4}{*}{D-CNN}& DC3-64-1& DC3-128-1 & DC3-256-1 & DC3-512-1 & DC3-512-2 &DC3-1024-12\\
			& C3-64-1& C3-128-1 & C3-256-1 & C3-512-1 & C3-512-2 &C1-1024-0\\
			& maxpool   & maxpool    & C3-256-1 & C3-512-1 & C3-512-2  &globalpool+concat\\
			&           &            & maxpool     & maxpool   & Davgpool   &\\
			\Xhline{3\arrayrulewidth}
		\end{tabular}
	\end{center}
	\caption{Network architecture. DeepLab is our baseline with a modified version of VGG-16 as the encoder. The convolution layer parameters are denoted as ``C[kernel size]-[number of channels]-[dilation]". ``DC" and ``Davgpool" represent depth-aware convolution and depth-aware average pooling respectively.}\vspace{-20pt}
	\label{table:netarch}
\end{table}

DeepLab\cite{deeplab} is a state-of-the-art method for semantic segmentation. We adopt DeepLab as our baseline for semantic segmentation and a modified VGG-16 network is used as the encoder. We replace layers in this network with depth-aware operations. The network configurations of the baseline and depth-aware CNN are outlined in Table~\ref{table:netarch}. Suppose $conv7$ has $C$ channels. Following~\cite{xiaojuaniccv17}, global pooling is used to compute a $C$-dim vector from $conv7$. This vector is then appended to all spatial positions and results in a $2C$-channel feature map. This feature map is followed by a $1\times 1$ conv layer and produce the segmentation probability map.

%------------------------------------------------------------------------
\section{Experiments}

Evaluation is performed on three popular RGB-D datasets: 
\begin{itemize}
\item NYUv2~\cite{nyuv2}: NYUv2 contains of $1,449$ RGB-D images with pixel-wise labels. We follow the $40$-class settings and the standard split with $795$ training images and $654$ testing images. 
\item SUN-RGBD~\cite{sunrgbd,Janoch2011dataset}: This dataset have $37$ categories of objects and consists of $10,335$ RGB-D images, with $5,285$ as training and $5050$ as testing. 
\item Stanford Indoor Dataset (SID)~\cite{2017arXiv170201105A}: SID contains $70,496$ RGB-D images with $13$ object categories. We use Area $1,2,3,4$ and $6$ as training, and Area $5$ as testing.
\end{itemize}

Four common metrics are used for evaluation: pixel accuracy (Acc), mean pixel accuracy of different categories (mAcc), mean Intersection-over-Union of different categories (mIoU), and frequency-weighted IoU (fwIoU). Suppose $n_{ij}$ is the number of pixels with ground truth class $i$ and predicted as class $j$, $n_C$ is the number of classes and $s_i$ is the number of pixels with ground truth class $i$, the total number of all pixels is $s = \sum_{i}s_i$. The four metrics are defined as follows: Acc = $\sum_{i}\frac{n_{ii}} { s}$, mAcc = $\frac{1}{n_C}\sum_{i}\frac{n_{ii}} {s_i}$, mIoU = $\frac{1}{n_C}\sum_{i}\frac{n_{ii}} {s_i+\sum_{j}n_{ji}-n_{ii}}$, fwIoU = $\frac{1}{s}\sum_{i}s_i\frac{n_{ii}} {s_i+\sum_{j}n_{ji}-n_{ii}}$.
%\begin{itemize}
%	\item Acc = $\sum_{i}\frac{n_{ii}} { s}$
%	\item mAcc = $\frac{1}{n_C}\sum_{i}\frac{n_{ii}} {s_i}$
%	\item mIoU = $\frac{1}{n_C}\sum_{i}\frac{n_{ii}} {s_i+\sum_{j}n_{ji}-n_{ii}}$
%	\item fwIoU = $\frac{1}{s}\sum_{i}s_i\frac{n_{ii}} {s_i+\sum_{j}n_{ji}-n_{ii}}$
%\end{itemize}

\paragraph{\bfseries{Implementation Details}}
For most experiments, DeepLab with a modified VGG-16 encoder (c.f. Table~\ref{table:netarch}) is the baseline. Depth-aware CNN based on DeepLab outlined in Table~\ref{table:netarch} is evaluated to validate the effectiveness of our approach and this is referred as ``D-CNN" in the paper. We also conduct experiments with combining HHA encoding~\cite{eccv14hha}. Following \cite{fcn,xiaojuaniccv17,eigeniccv15}, two baseline networks consume RGB and HHA images separately and the predictions of both networks are summed up in the last layer. This two-stream network is dubbed as ``HHA". To make fair comparison, we also build depth-aware CNN with this two-stream fashion and denote this as ``D-CNN+HHA".
In ablation study, we further replace VGG-16 with ResNet-50~\cite{resnet} as the encoder to have a better understanding of the functionality of depth-aware operations.

We use SGD optimizer with initial learning rate $0.001$, momentum $0.9$ and batch size $1$. The learning rate is multiplied by $(1-\frac{iter}{max\_iter})^{0.9}$ for every $10$ iterarions. $\alpha$ is set to $8.3$. (The impact of $\alpha$ is studied in Section~\ref{sec:ablation}.) The dataset is augmented by randomly scaling, cropping, and color jittering. We use PyTorch deep learning framework. Both depth-aware convolution and depth-aware average pooling operators are implemented with CUDA acceleration. Code will be released.
\vspace{-10pt}

\subsection{Main Results}
\label{sec:mainresults}
Depth-aware CNN is compared with both its baseline and the state-of-the-art methods on NYUv2 and SUN-RGBD dataset. It is also compared with the baseline on SID dataset.
\paragraph{\bfseries{NYUv2}}
Table~\ref{table:nyud2scratch} shows quantitative comparison results between D-CNNs and baseline models. Since D-CNN and its baseline are in different function space, all networks are trained from scratch to make fair comparison in this experiment. Without introducing any parameters, D-CNN outperforms the baseline by incorporating geometric information in convolution operation. Moreover, the performance of D-CNN also exceeds ``HHA" network by using only half of its parameters. This effectively validate the superior capability of D-CNN on handling geometry over ``HHA".  
\vspace{-10pt}
\begin{table}
\begin{center}
\newcolumntype{C}{>{\centering\arraybackslash}p{3.5em}}
\newcolumntype{E}{>{\centering\arraybackslash}p{7em}}
\begin{tabular}{c|ECCE}
	\Xhline{3\arrayrulewidth}
& Baseline& HHA & D-CNN & \small{D-CNN+HHA}\\
\hline
Acc (\%)&50.1&59.1&60.3 &\bf{61.4}\\
mAcc (\%)&23.9&30.8&\bf{39.3} &35.6\\
mIoU (\%)&15.9&21.9&\bf{27.8} & 26.2\\
fwIoU (\%)&34.2&43.0&44.9 &\bf{45.7}\\
\Xhline{3\arrayrulewidth}
\end{tabular}
\end{center}
\caption{Comparison with baseline CNNs on NYUv2 test set. Networks are trained from scratch.}\vspace{-40pt}
\label{table:nyud2scratch}
\end{table}

\begin{table}
	\begin{center}
		\newcolumntype{C}{>{\centering\arraybackslash}p{3.5em}}
		\newcolumntype{D}{>{\centering\arraybackslash}p{2.6em}}
		\newcolumntype{E}{>{\centering\arraybackslash}p{7em}}
		\begin{tabular}{c|DDDD|CCE}
			\Xhline{3\arrayrulewidth}
			&\cite{fcn}&\cite{eigeniccv15}&\cite{yang_cvpr17}&\cite{xiaojuaniccv17}& \small{HHA} & D-CNN & \small{D-CNN+HHA}\\
			\hline
			mAcc (\%)& 46.1 & 45.1&53.8 & 55.2 &51.1&53.6&\bf{56.3}\\
			mIoU (\%)& 34.0 & 34.1&40.1 & 42.0 &40.4 &41.0&\bf{43.9}\\
			\Xhline{3\arrayrulewidth}
		\end{tabular}
	\end{center}
	\caption{Comparison with the state-of-the-arts on NYUv2 test set. Networks are trained from pre-trained models.}\vspace{-20pt}
	\label{table:nyud2}
\end{table}
We also compare our results with the state-of-the-art methods. Table~\ref{table:nyud2} illustrates the good performance of D-CNN. In this experiment, the networks are initialized with the pre-trained parameters in \cite{deeplab}. Long et al.~\cite{fcn} and Eigen et al.~\cite{eigeniccv15} both use the two-stream network with HHA/depth encoding. Yang et al.~\cite{yang_cvpr17} compute optical flows and superpixels to augment the performance with spatial-temporal information. D-CNN with only one VGG network is superior to their methods. Qi et al.~\cite{xiaojuaniccv17} built a 3D graph on the top of VGG encoder and use RNN to update the graph, which introduces more network parameters and higher computation complexity. As is shown in Table~\ref{table:nyud2}, D-CNN is already comparable with these state-of-the-art methods. By incorporating HHA encoding, our method achieves the state-of-the-art on this dataset. Figure~\ref{fig:nyud2} visualizes qualitative comparison results on NYUv2 test set.

\begin{figure}
	\centering
	\newcolumntype{C}{>{\centering\arraybackslash}p{5em}}
	\begin{tabular}{CCCCCCC}
		\includegraphics[width=0.13\textwidth]{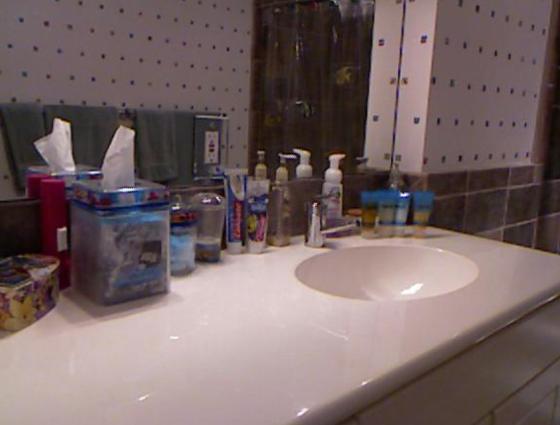}
		&\includegraphics[width=0.13\textwidth]{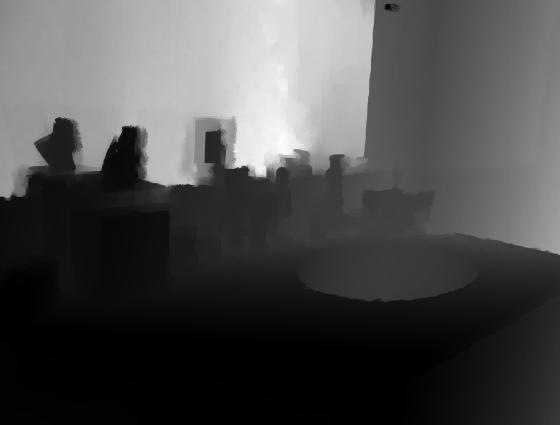}
		&\includegraphics[width=0.13\textwidth]{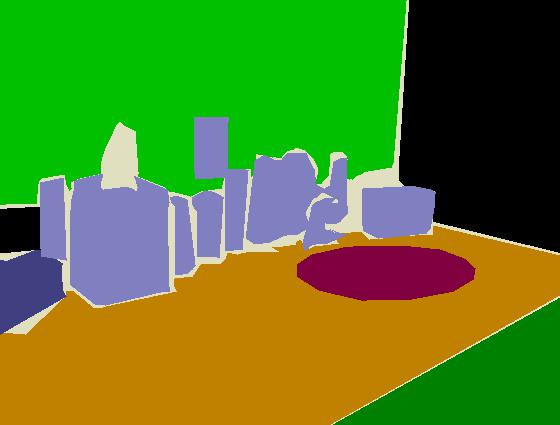}
		&\includegraphics[width=0.13\textwidth]{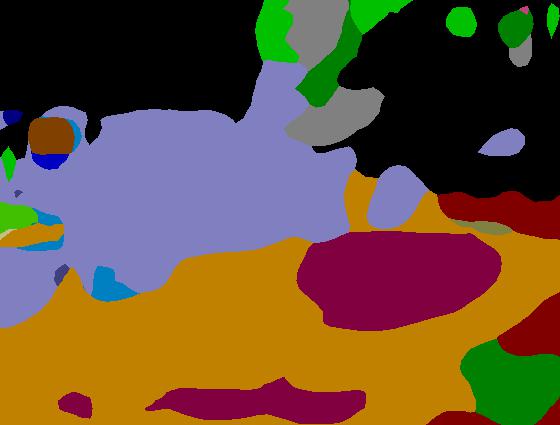}
		&\includegraphics[width=0.13\textwidth]{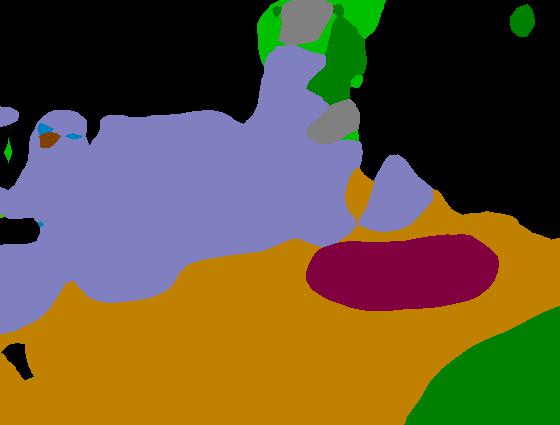}
		&\includegraphics[width=0.13\textwidth]{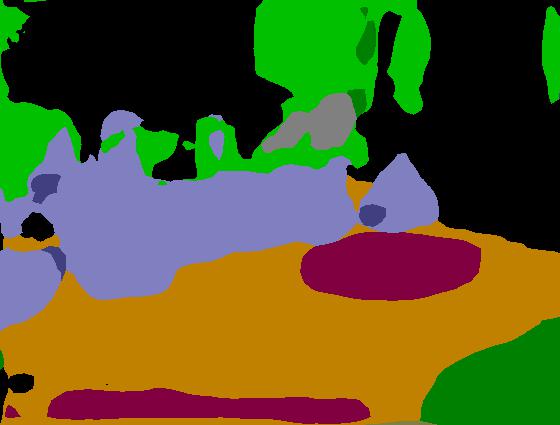}
		&\includegraphics[width=0.13\textwidth]{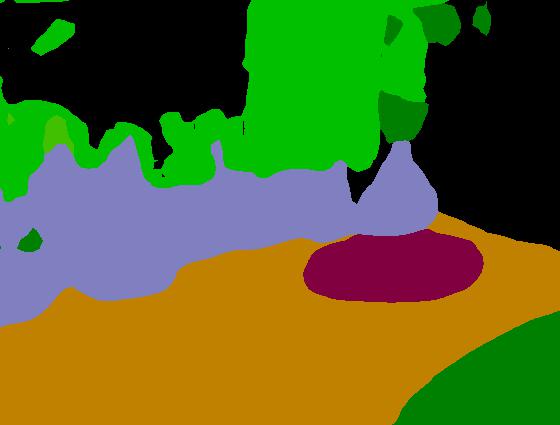}
		\\
		\includegraphics[width=0.13\textwidth]{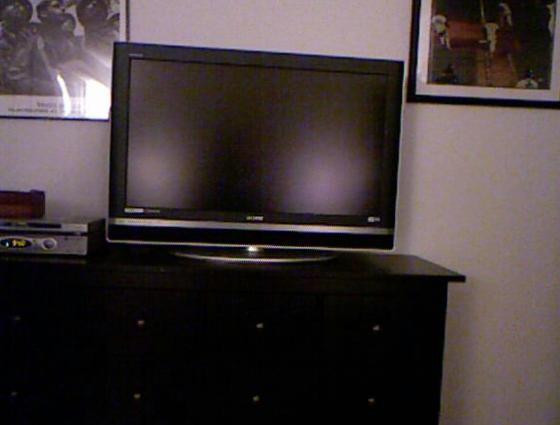}
		&\includegraphics[width=0.13\textwidth]{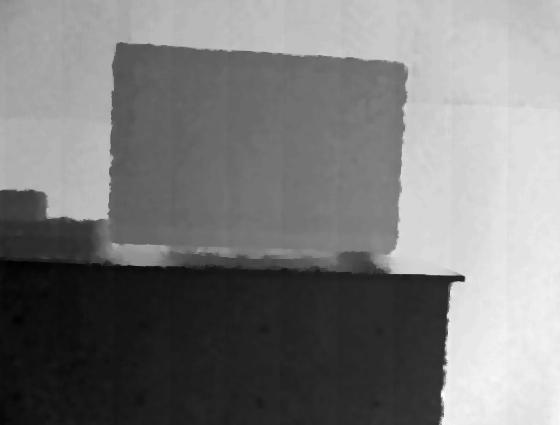}
		&\includegraphics[width=0.13\textwidth]{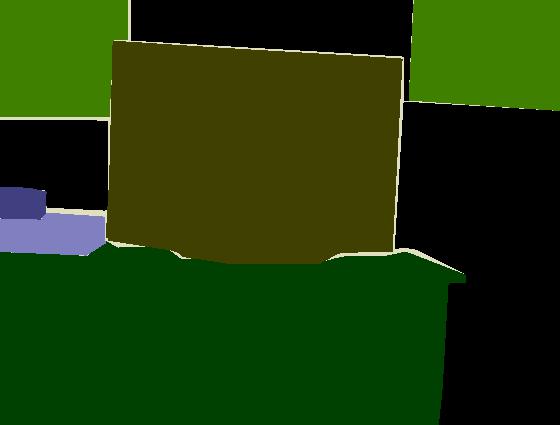}
		&\includegraphics[width=0.13\textwidth]{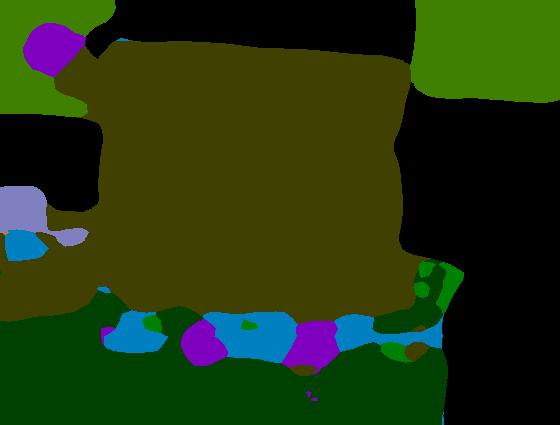}
		&\includegraphics[width=0.13\textwidth]{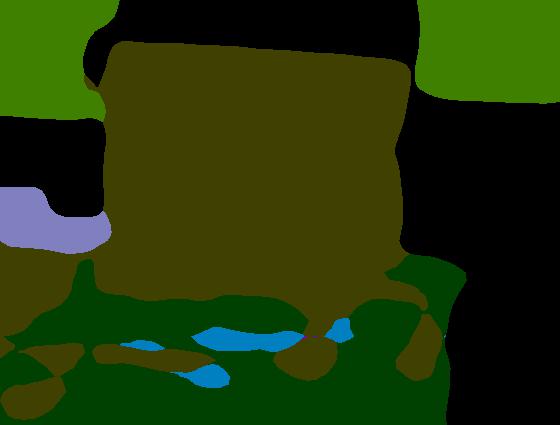}
		&\includegraphics[width=0.13\textwidth]{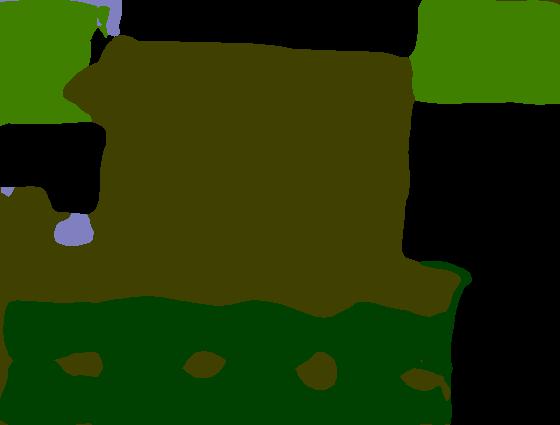}
		&\includegraphics[width=0.13\textwidth]{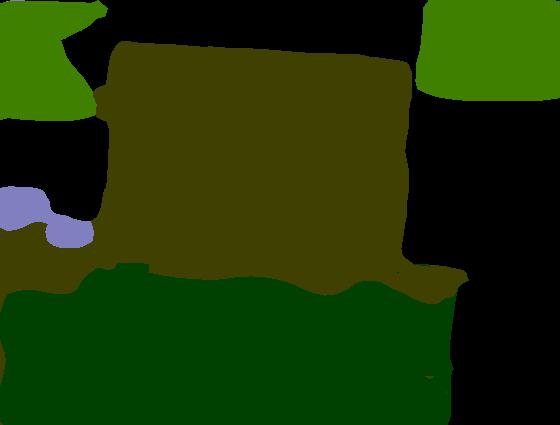}
		\\
		\includegraphics[width=0.13\textwidth]{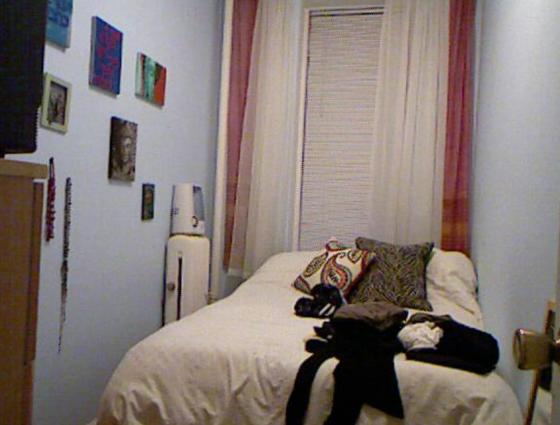}
		&\includegraphics[width=0.13\textwidth]{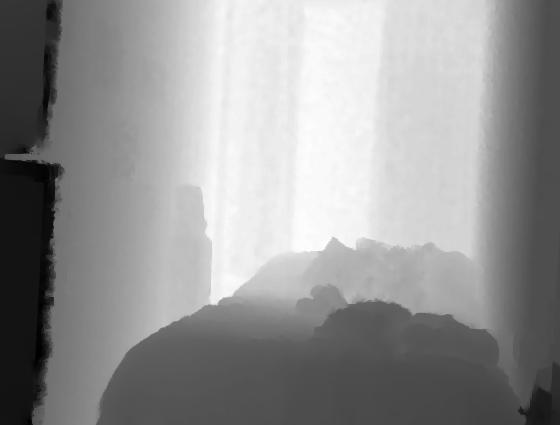}
		&\includegraphics[width=0.13\textwidth]{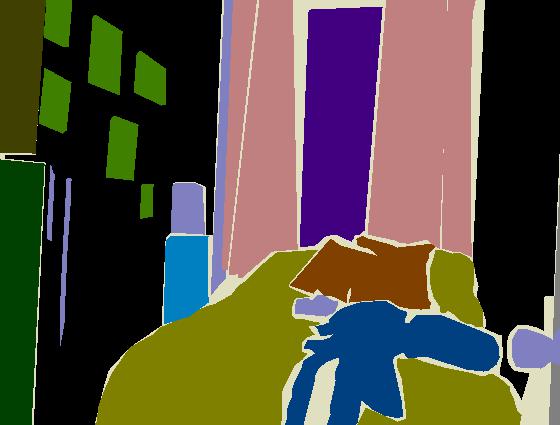}
		&\includegraphics[width=0.13\textwidth]{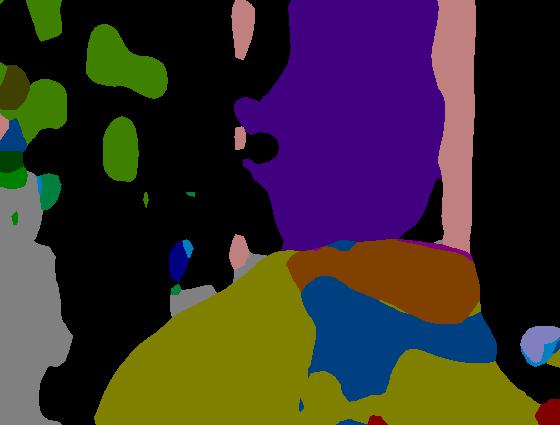}
		&\includegraphics[width=0.13\textwidth]{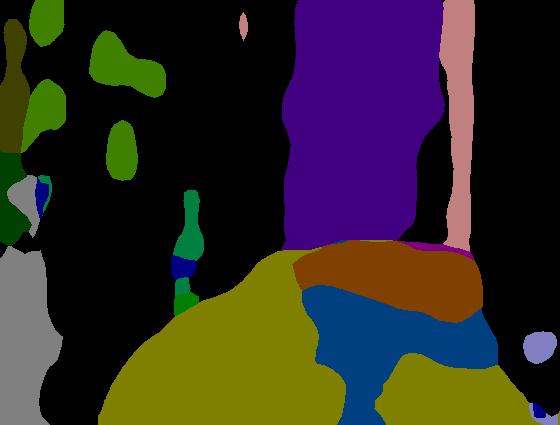}
		&\includegraphics[width=0.13\textwidth]{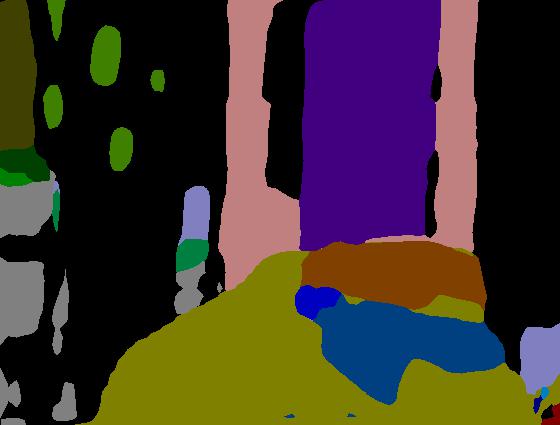}
		&\includegraphics[width=0.13\textwidth]{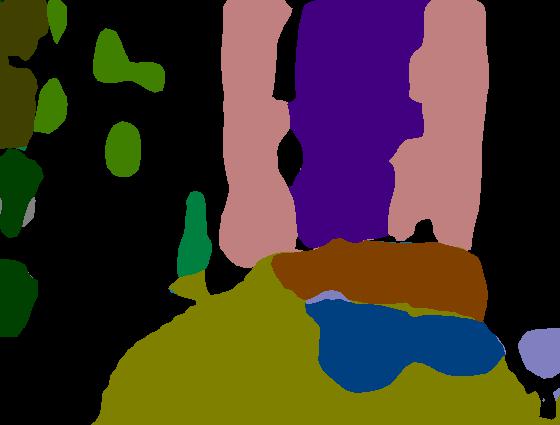}
		\\
		\includegraphics[width=0.13\textwidth]{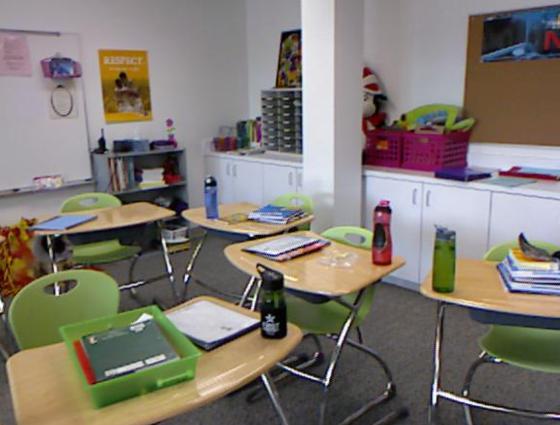}
		&\includegraphics[width=0.13\textwidth]{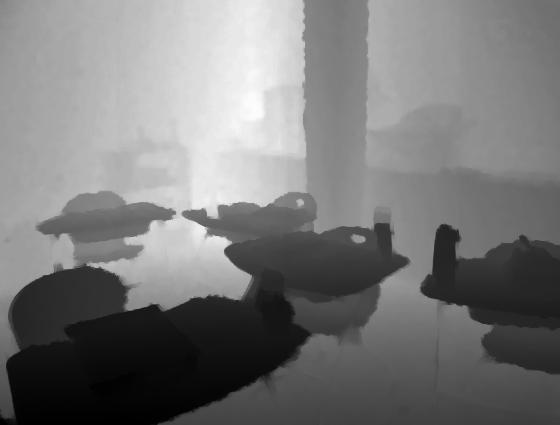}
		&\includegraphics[width=0.13\textwidth]{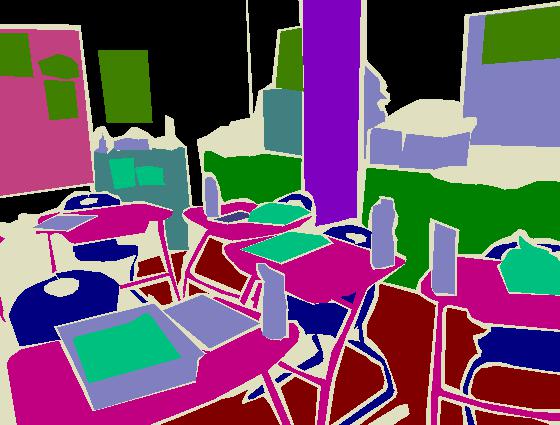}
		&\includegraphics[width=0.13\textwidth]{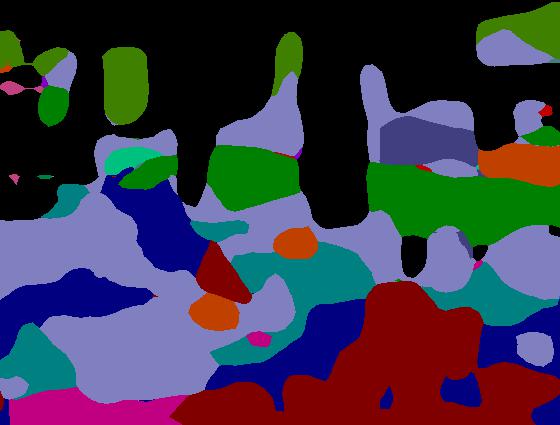}
		&\includegraphics[width=0.13\textwidth]{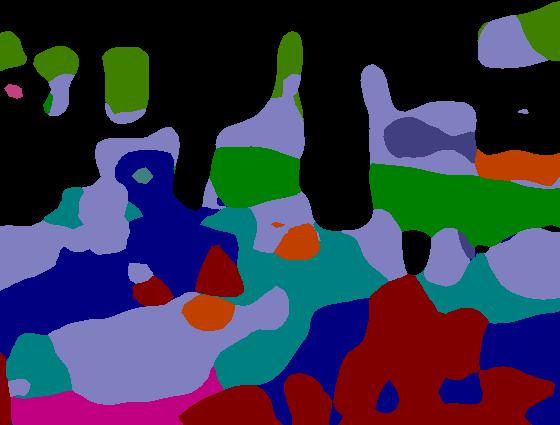}
		&\includegraphics[width=0.13\textwidth]{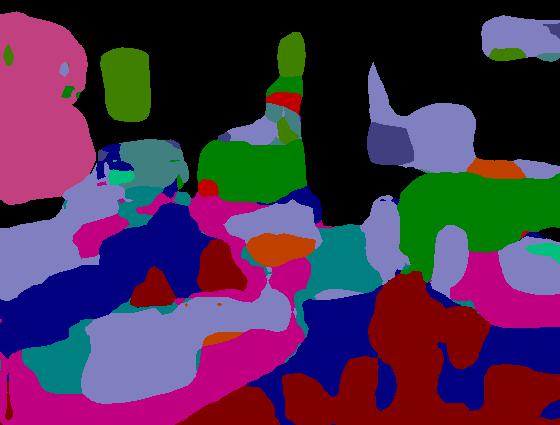}
		&\includegraphics[width=0.13\textwidth]{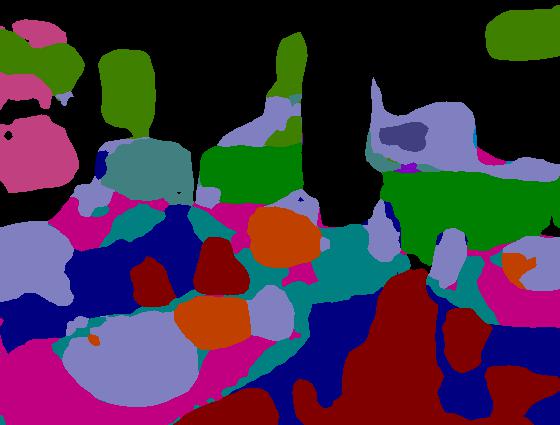}
		\\
		\includegraphics[width=0.13\textwidth]{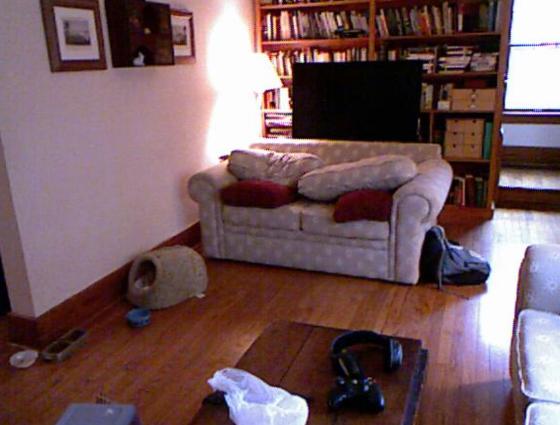}
		&\includegraphics[width=0.13\textwidth]{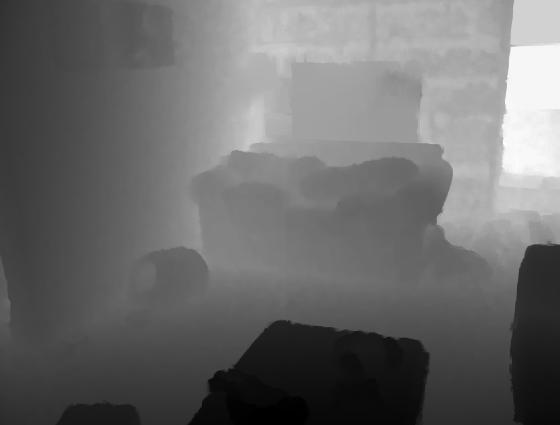}
		&\includegraphics[width=0.13\textwidth]{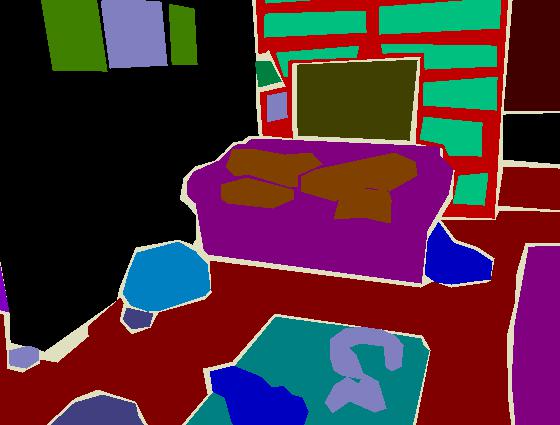}
		&\includegraphics[width=0.13\textwidth]{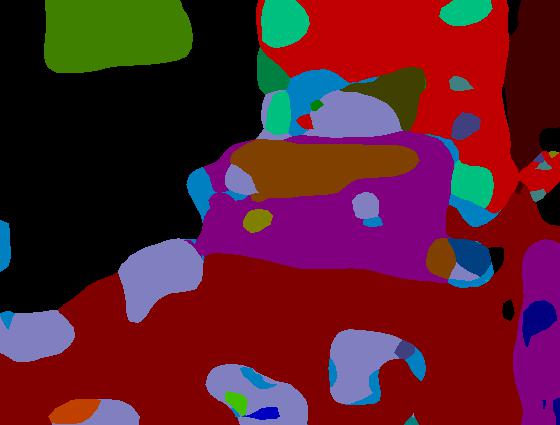}
		&\includegraphics[width=0.13\textwidth]{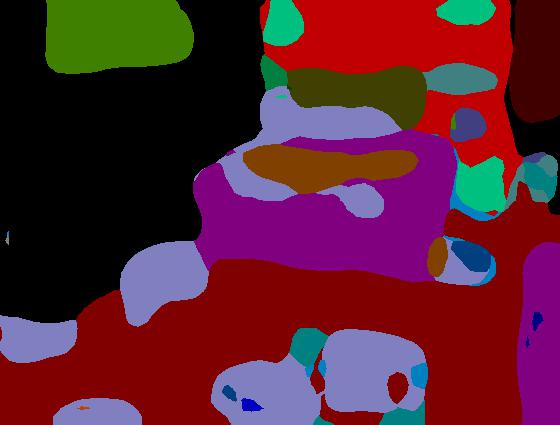}
		&\includegraphics[width=0.13\textwidth]{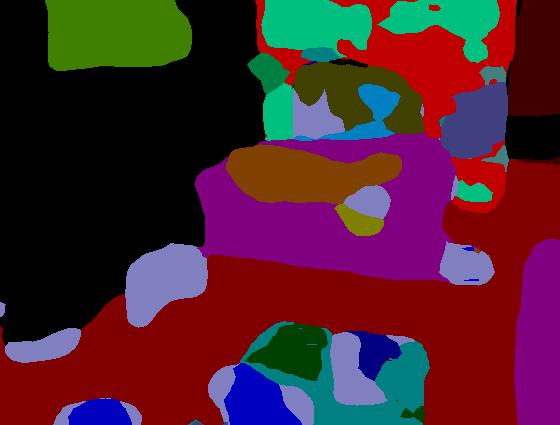}
		&\includegraphics[width=0.13\textwidth]{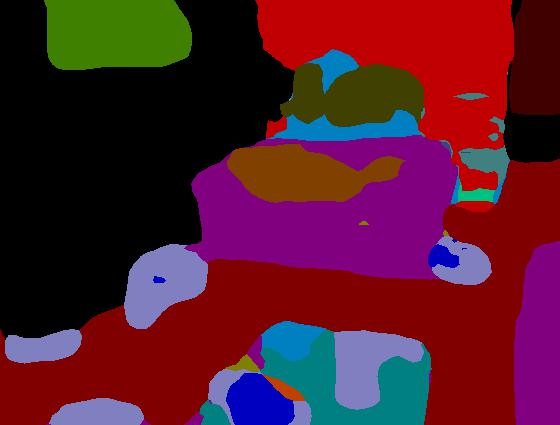}
		\\
		\includegraphics[width=0.13\textwidth]{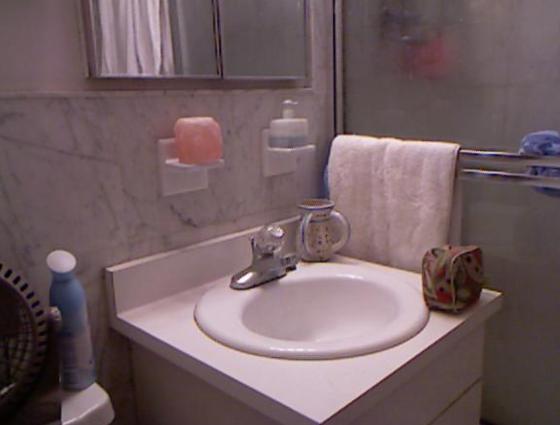}
		&\includegraphics[width=0.13\textwidth]{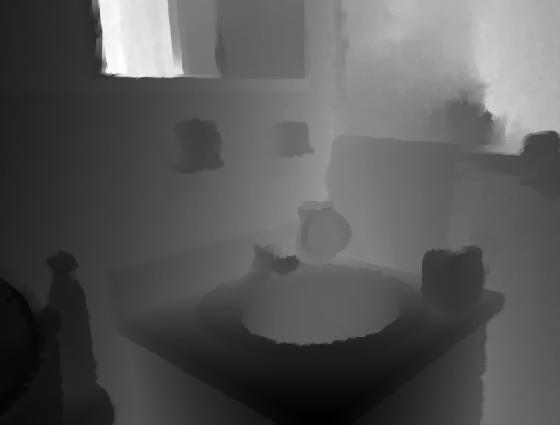}
		&\includegraphics[width=0.13\textwidth]{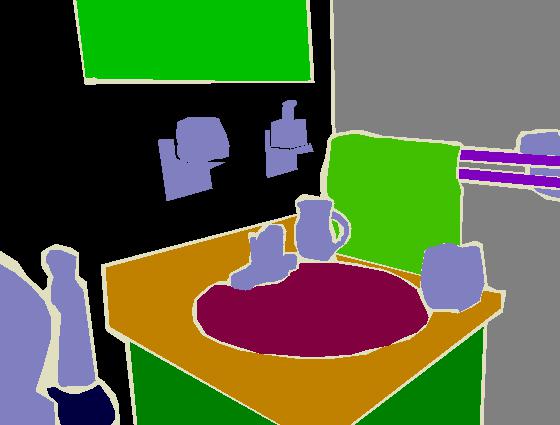}
		&\includegraphics[width=0.13\textwidth]{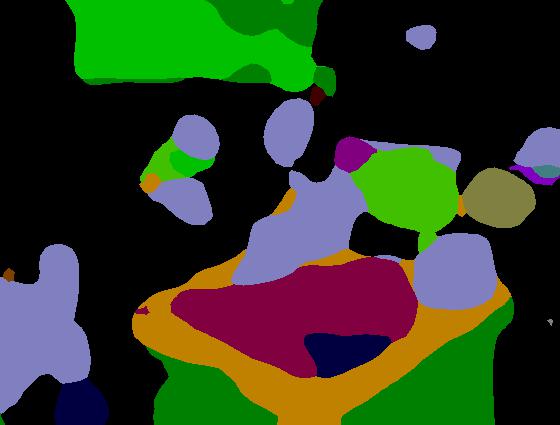}
		&\includegraphics[width=0.13\textwidth]{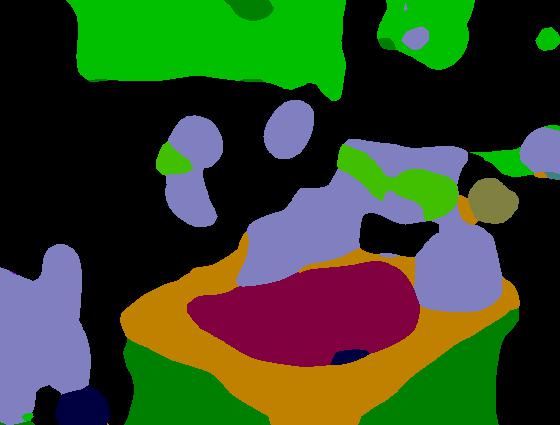}
		&\includegraphics[width=0.13\textwidth]{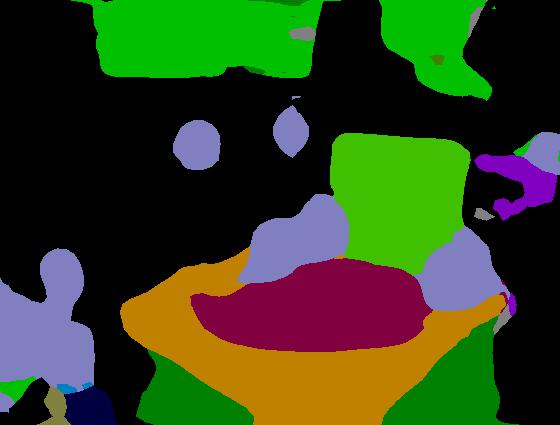}
		&\includegraphics[width=0.13\textwidth]{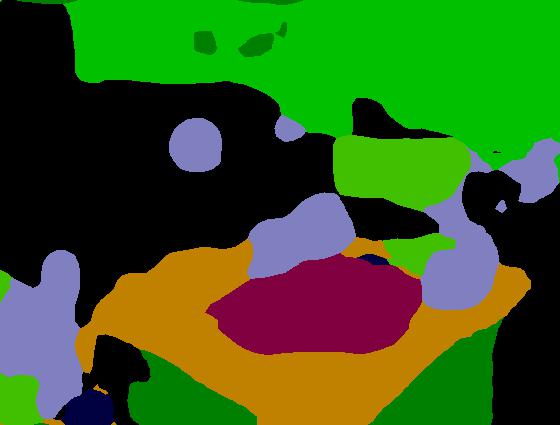}
		\\
		\includegraphics[width=0.13\textwidth]{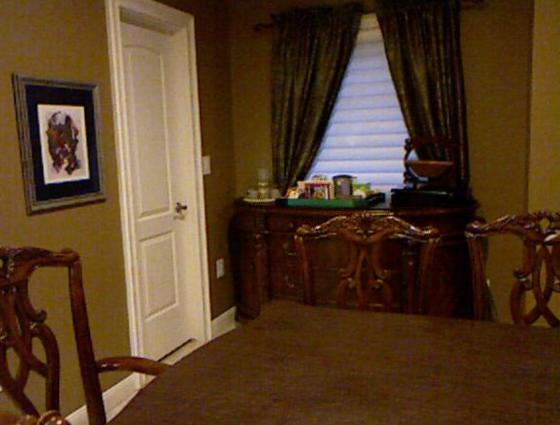}
		&\includegraphics[width=0.13\textwidth]{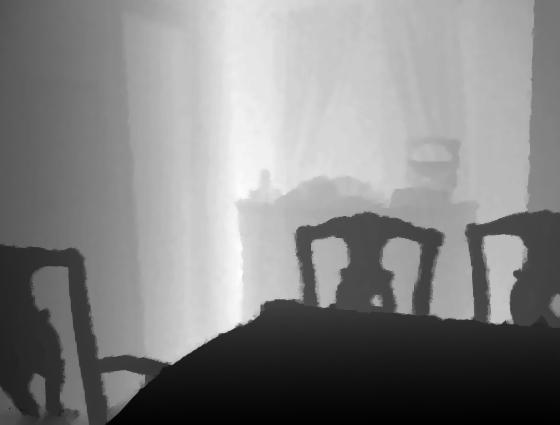}
		&\includegraphics[width=0.13\textwidth]{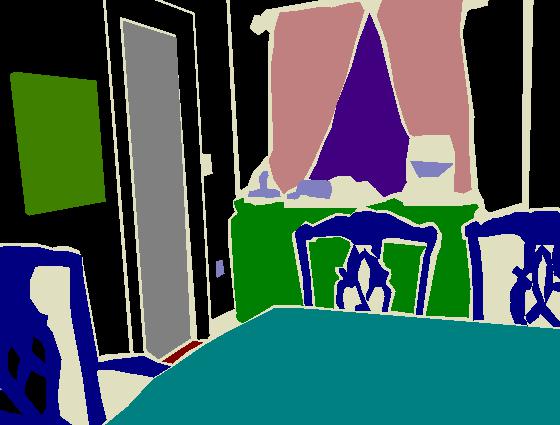}
		&\includegraphics[width=0.13\textwidth]{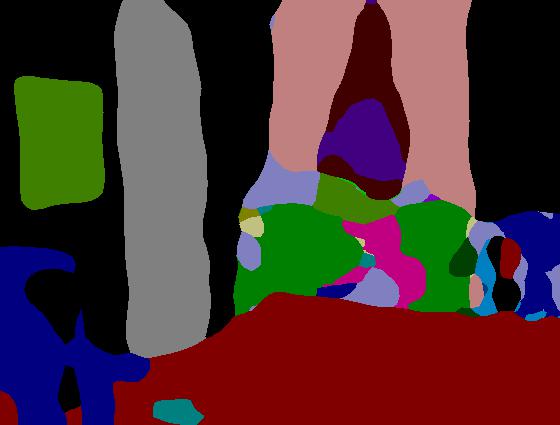}
		&\includegraphics[width=0.13\textwidth]{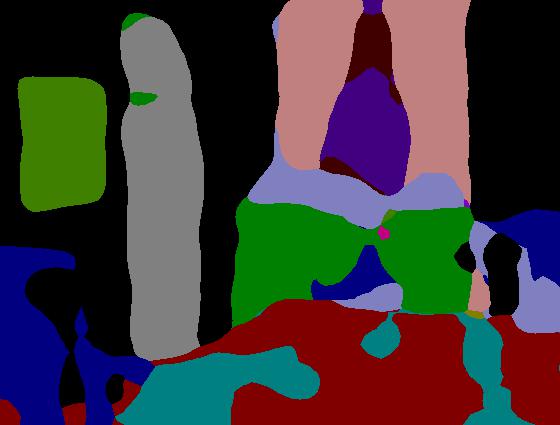}
		&\includegraphics[width=0.13\textwidth]{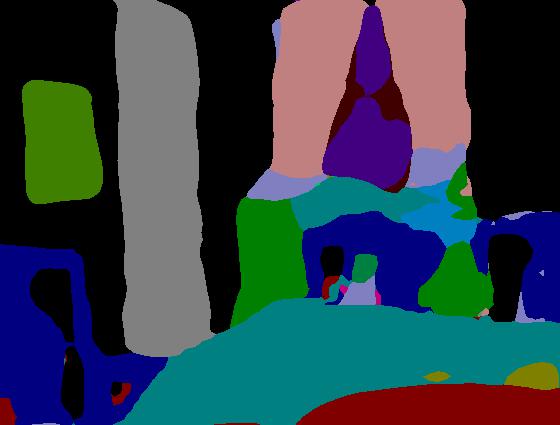}
		&\includegraphics[width=0.13\textwidth]{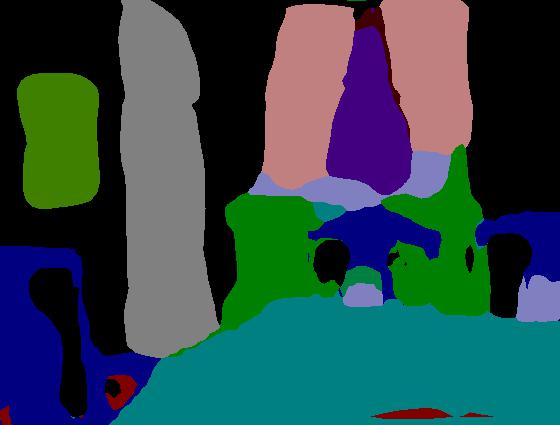}
		\\
		\includegraphics[width=0.13\textwidth]{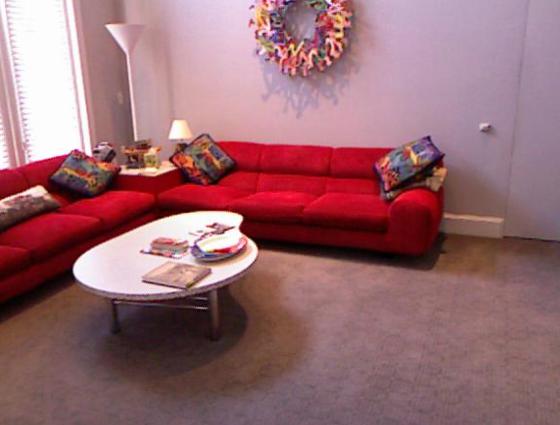}
		&\includegraphics[width=0.13\textwidth]{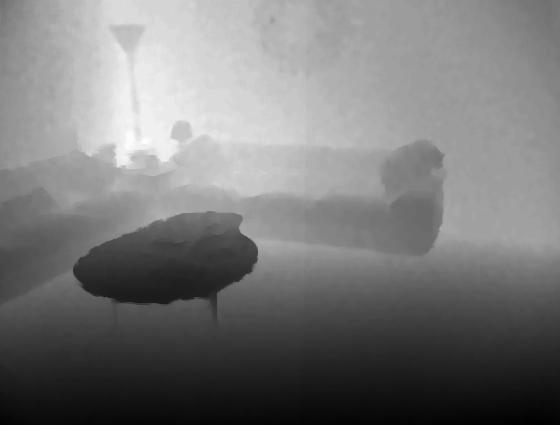}
		&\includegraphics[width=0.13\textwidth]{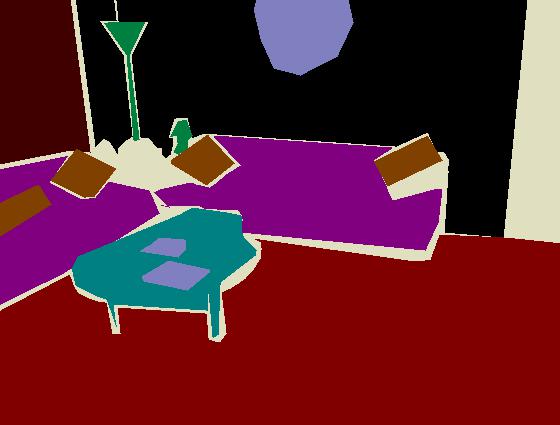}
		&\includegraphics[width=0.13\textwidth]{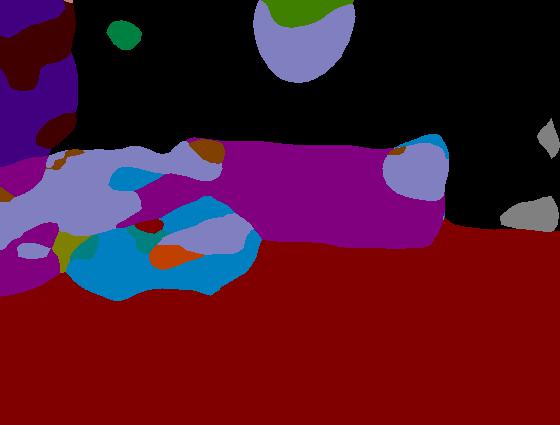}
		&\includegraphics[width=0.13\textwidth]{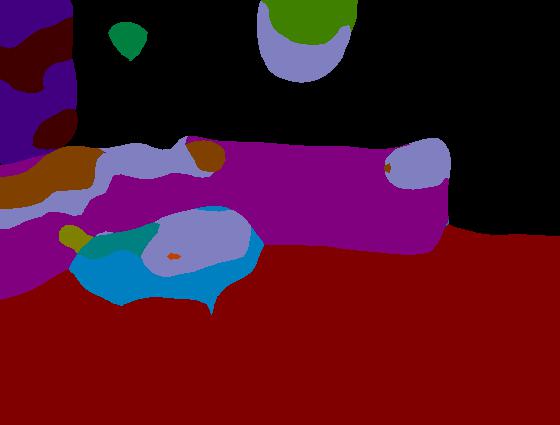}
		&\includegraphics[width=0.13\textwidth]{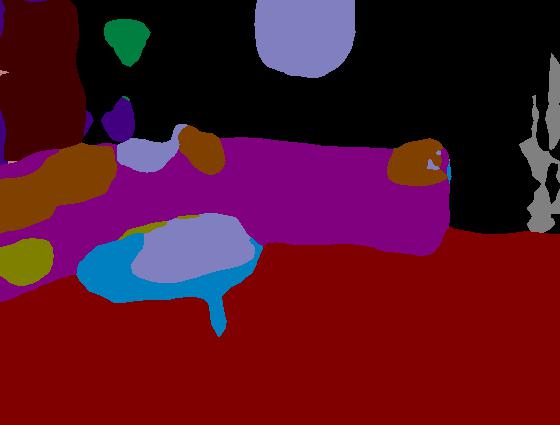}
		&\includegraphics[width=0.13\textwidth]{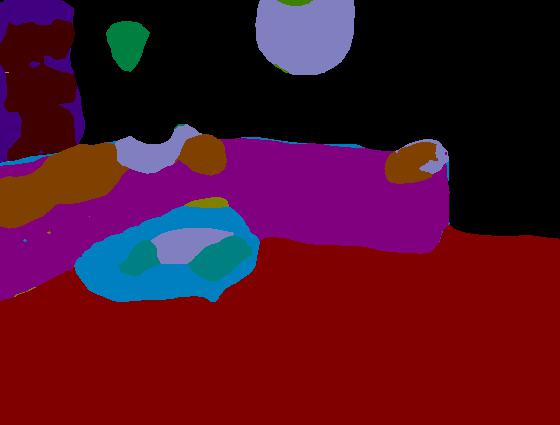}
		\\
		\tiny{RGB}& \tiny{Depth}& \tiny{GT}& \tiny{Baseline} & \tiny{HHA }&\tiny{D-CNN}& \tiny{DCNN+HHA}
	\end{tabular}
	\caption{Segmentation results on NYUv2 test dataset. ``GT" denotes ground truth. The white regions in ``GT" are the ignoring category. Networks are trained from pre-trained models.}\vspace{-20pt}
	\label{fig:nyud2}
\end{figure}

\paragraph{\bfseries{SUN-RGBD}}
The comparison results between D-CNN and its baseline are listed in Table~\ref{table:sunrgbdscratch}. The networks in this table are trained from scratch. D-CNN outperforms baseline by a large margin. Substituting the baseline with the two-stream ``HHA" network is able to further improve the performance.

By comparing with the state-of-the-art methods in Table~\ref{table:sunrgbd}, we can further see the effectiveness of D-CNN. Similarly as in NYUv2, the networks are initialized with pre-trained model in this experiment.  Figure~\ref{fig:sunrgbd} illustrates the qualitative comparison results on SUN-RGBD test set. Our network achieves comparable performance with the state-of-the-art method~\cite{xiaojuaniccv17}, while their method is more time-consuming. We will further compare the runtime and numbers of model parameters in Section~\ref{sec:time}. 
\vspace{-10pt}
\begin{table}
\begin{center}
\newcolumntype{C}{>{\centering\arraybackslash}p{3.5em}}
\newcolumntype{E}{>{\centering\arraybackslash}p{7em}}
\begin{tabular}{c|ECCE}
	\Xhline{3\arrayrulewidth}
& Baseline& HHA & D-CNN & \small{D-CNN+HHA}\\
\hline
Acc (\%)& 66.6&72.6&72.4&\bf{72.9}\\
mAcc (\%)& 31.5&37.9&38.6&\bf{41.2}\\
mIoU (\%)& 22.8&28.8&29.7&\bf{31.3}\\
fwIoU (\%)& 51.4&58.5&58.2&\bf{59.3}\\
\Xhline{3\arrayrulewidth}
\end{tabular}
\end{center}
\caption{Comparison with baseline CNNs on SUN-RGBD test set. Networks are trained from scratch.}\vspace{-10pt}
\label{table:sunrgbdscratch}
\end{table}

\begin{table}
\begin{center}
\newcolumntype{C}{>{\centering\arraybackslash}p{3.5em}}
\newcolumntype{D}{>{\centering\arraybackslash}p{2.6em}}
\newcolumntype{E}{>{\centering\arraybackslash}p{7em}}
\begin{tabular}{c|DD|CCE}
	\Xhline{3\arrayrulewidth}
&\cite{lstmcf}&\cite{xiaojuaniccv17}& HHA & D-CNN & D-CNN+HHA\\
\hline
mAcc (\%)& 48.1 & \bf{55.2}  & 50.5&51.2& 53.5\\
mIoU (\%)& - & \bf{42.0}  & 40.2&41.5& \bf{42.0}\\
\Xhline{3\arrayrulewidth}
\end{tabular}
\end{center}
\caption{Comparison with the state-of-the-arts on SUN-RGBD test set. Networks are trained from pre-trained models.}\vspace{-25pt}
\label{table:sunrgbd}
\end{table}

\begin{figure}[tb]
	\centering
	\newcolumntype{C}{>{\centering\arraybackslash}p{5em}}
	\begin{tabular}{CCCCCCC}
		\includegraphics[width=0.13\textwidth]{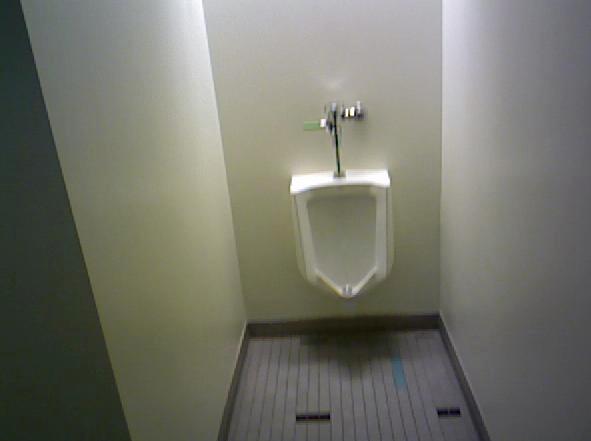}
		&\includegraphics[width=0.13\textwidth]{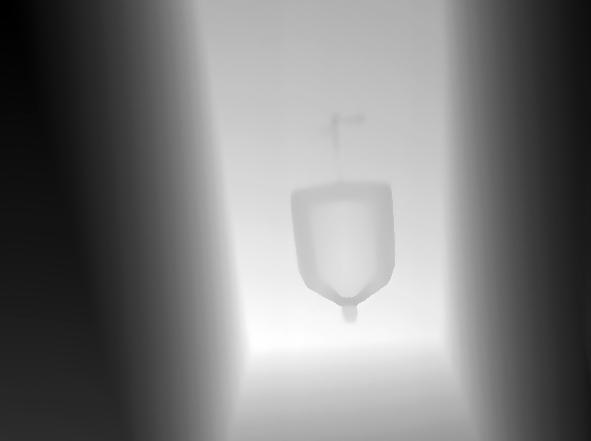}
		&\includegraphics[width=0.13\textwidth]{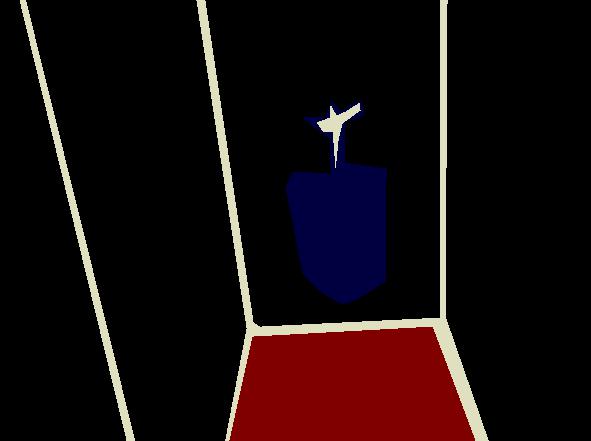}
		&\includegraphics[width=0.13\textwidth]{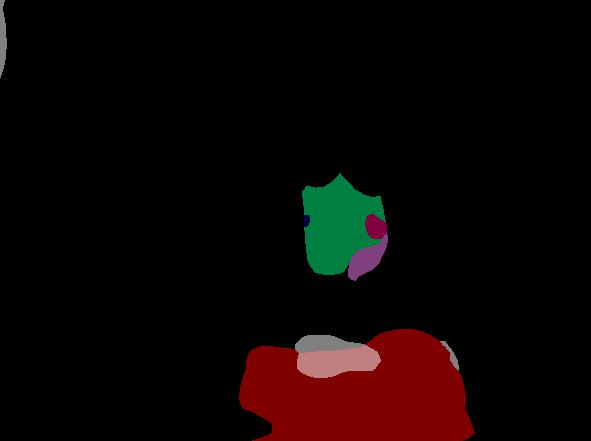}
		&\includegraphics[width=0.13\textwidth]{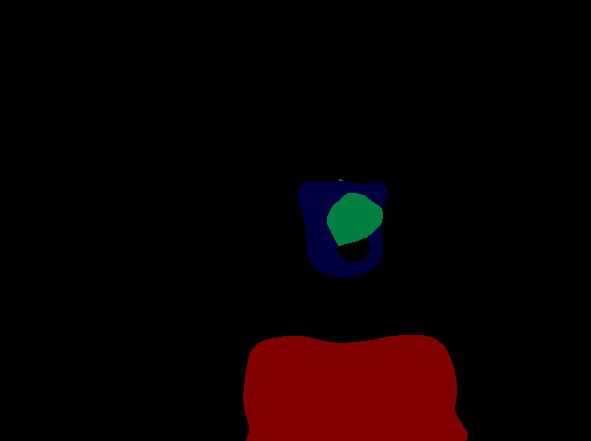}
		&\includegraphics[width=0.13\textwidth]{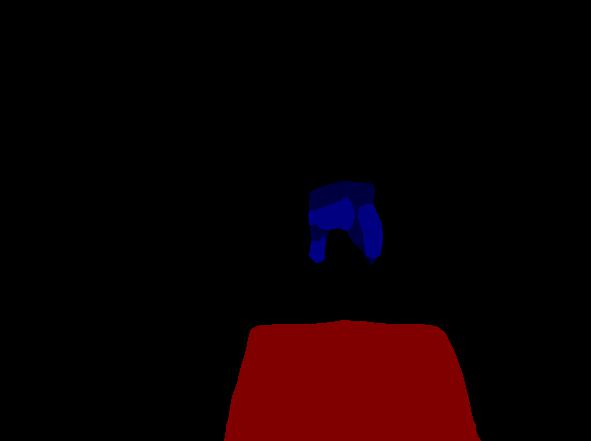}
		&\includegraphics[width=0.13\textwidth]{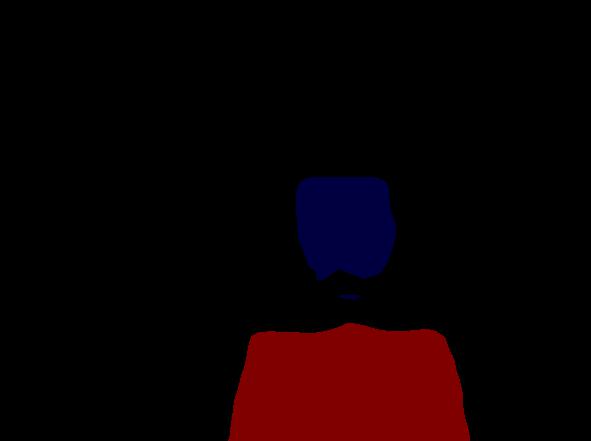}
		\\
		\includegraphics[width=0.13\textwidth]{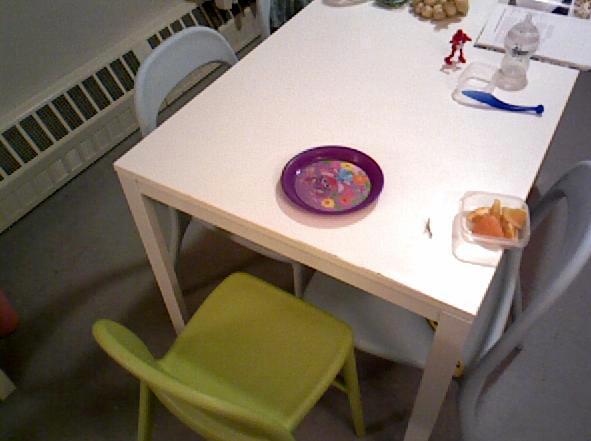}
		&\includegraphics[width=0.13\textwidth]{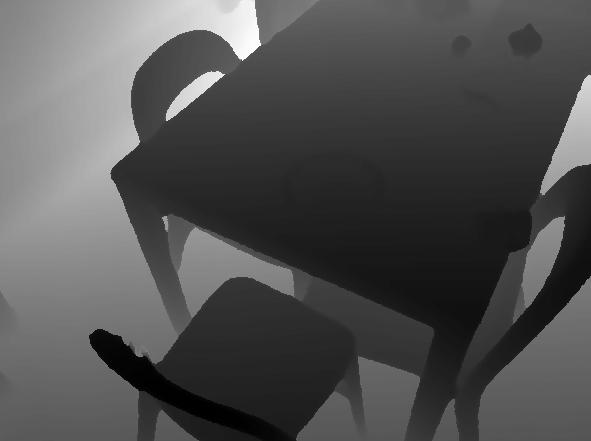}
		&\includegraphics[width=0.13\textwidth]{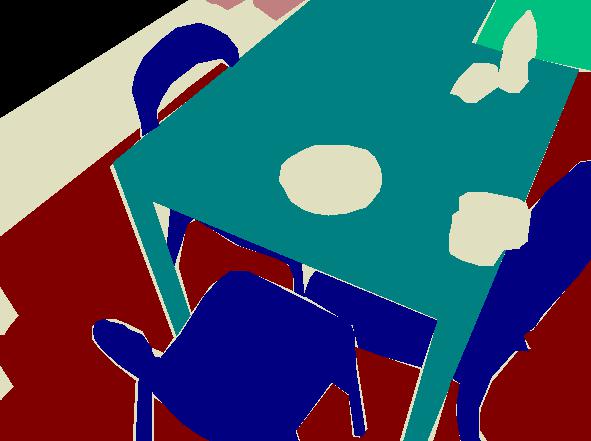}
		&\includegraphics[width=0.13\textwidth]{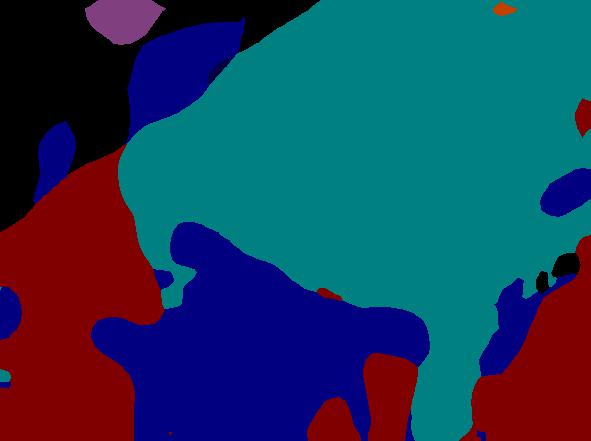}
		&\includegraphics[width=0.13\textwidth]{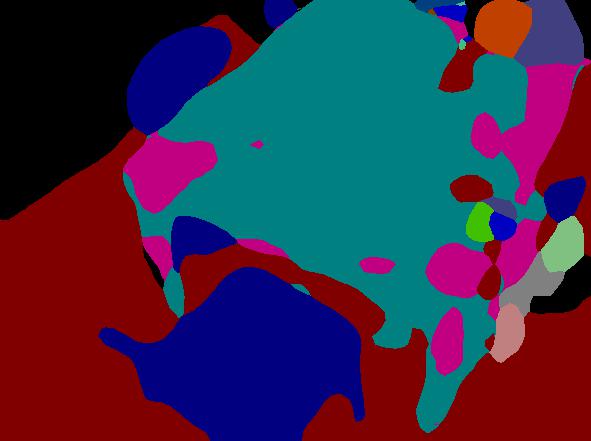}
		&\includegraphics[width=0.13\textwidth]{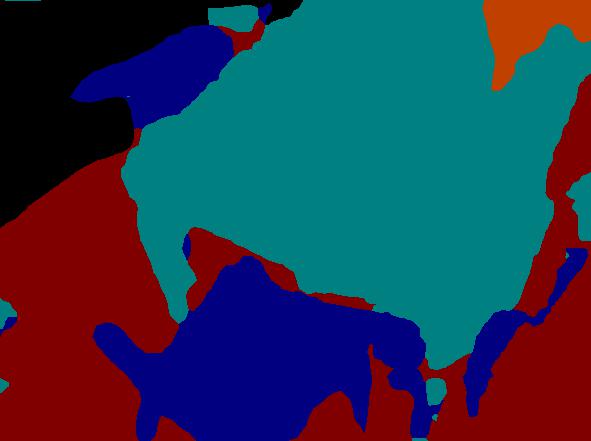}
		&\includegraphics[width=0.13\textwidth]{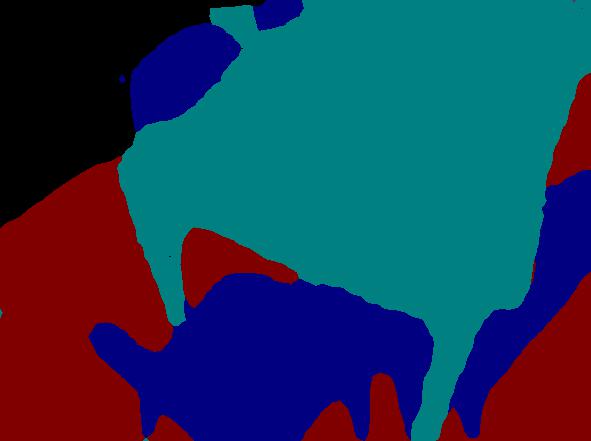}
		\\
		\includegraphics[width=0.13\textwidth]{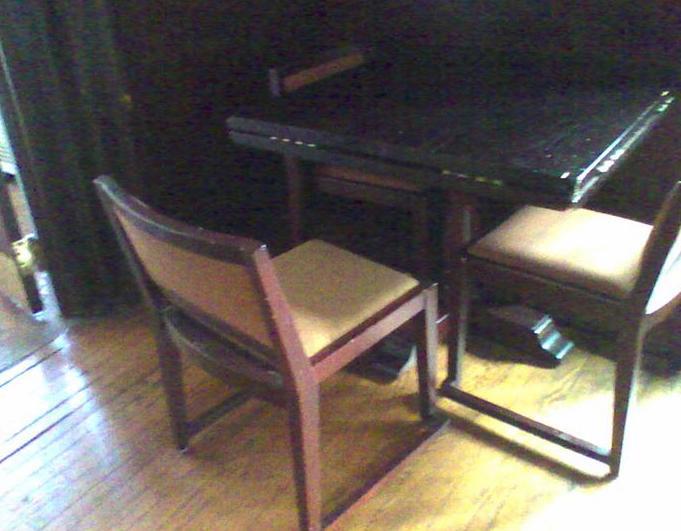}
		&\includegraphics[width=0.13\textwidth]{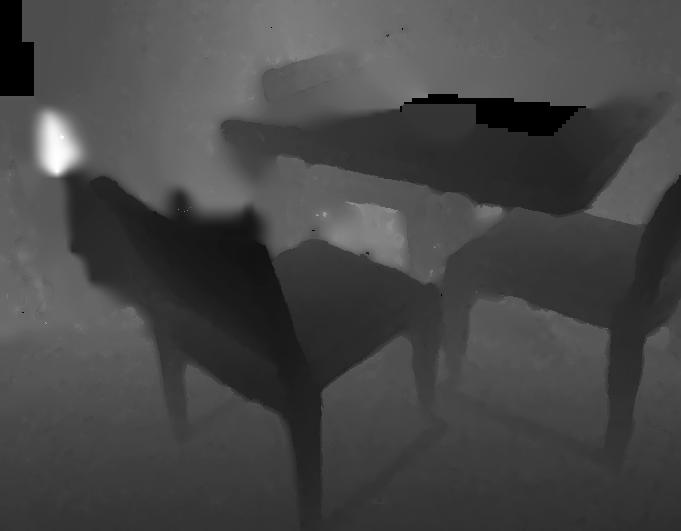}
		&\includegraphics[width=0.13\textwidth]{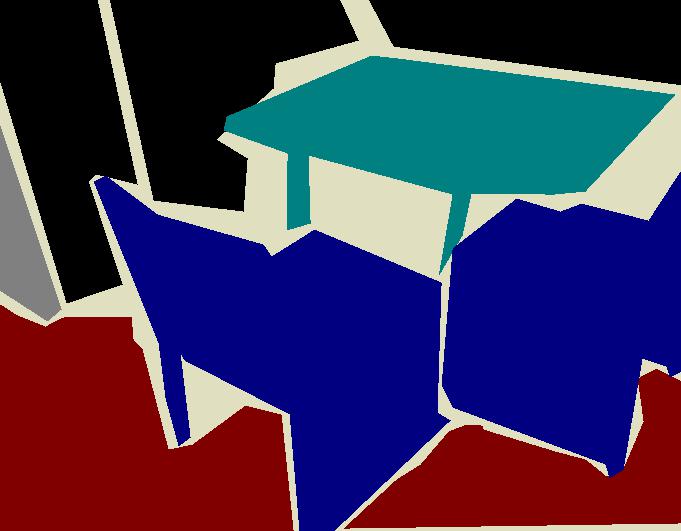}
		&\includegraphics[width=0.13\textwidth]{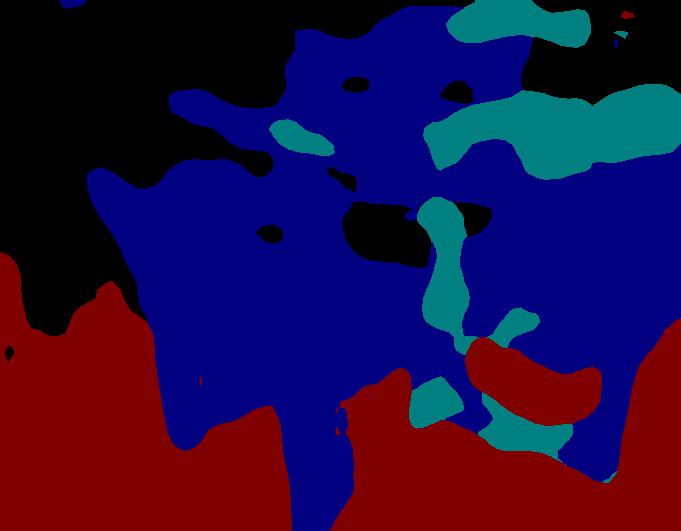}
		&\includegraphics[width=0.13\textwidth]{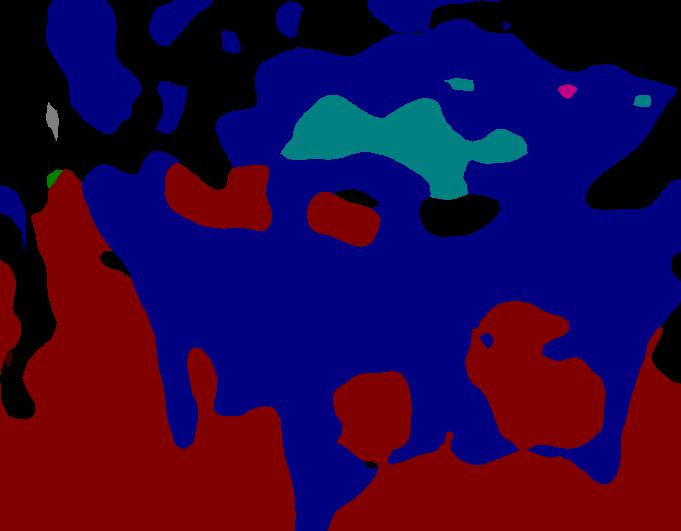}
		&\includegraphics[width=0.13\textwidth]{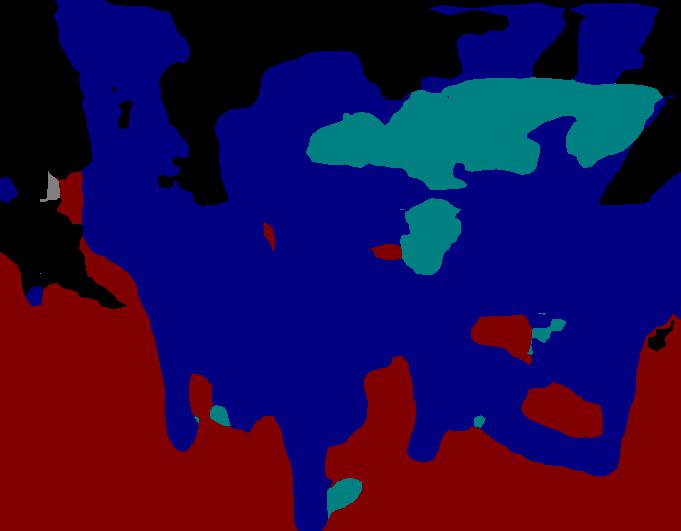}
		&\includegraphics[width=0.13\textwidth]{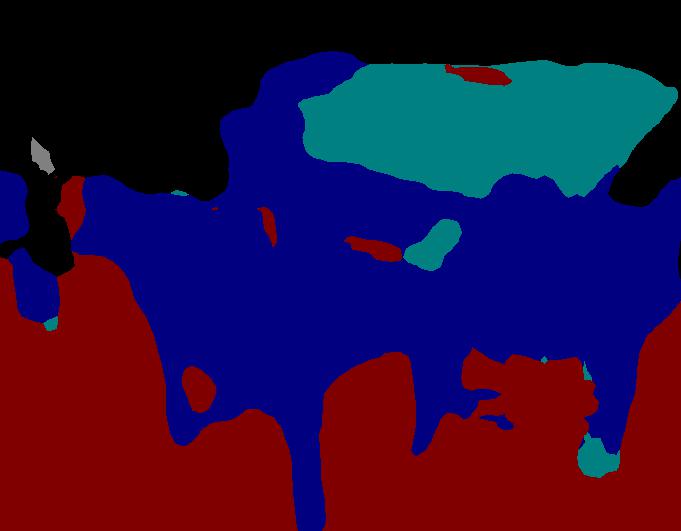}
		\\
		\includegraphics[width=0.13\textwidth]{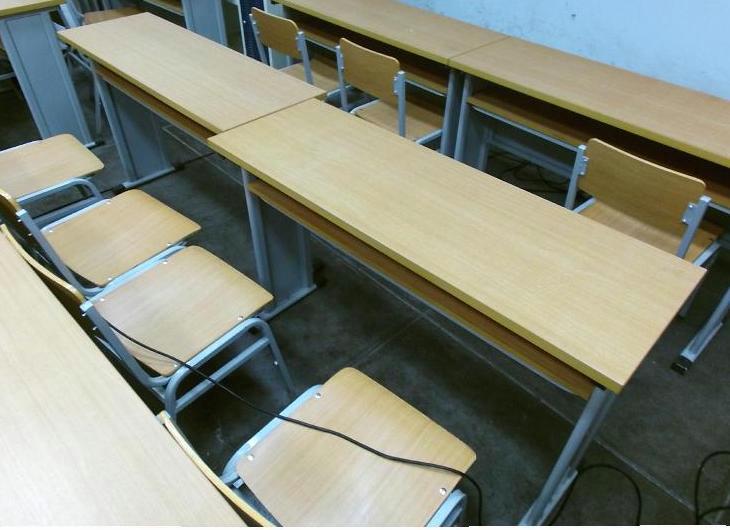}
		&\includegraphics[width=0.13\textwidth]{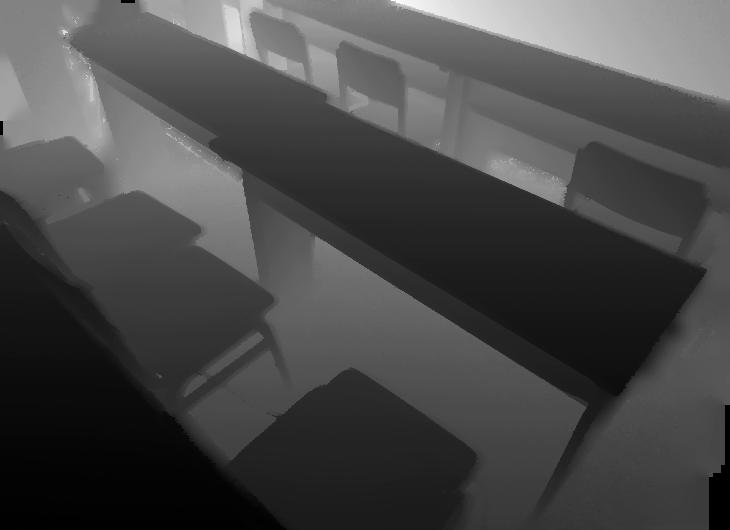}
		&\includegraphics[width=0.13\textwidth]{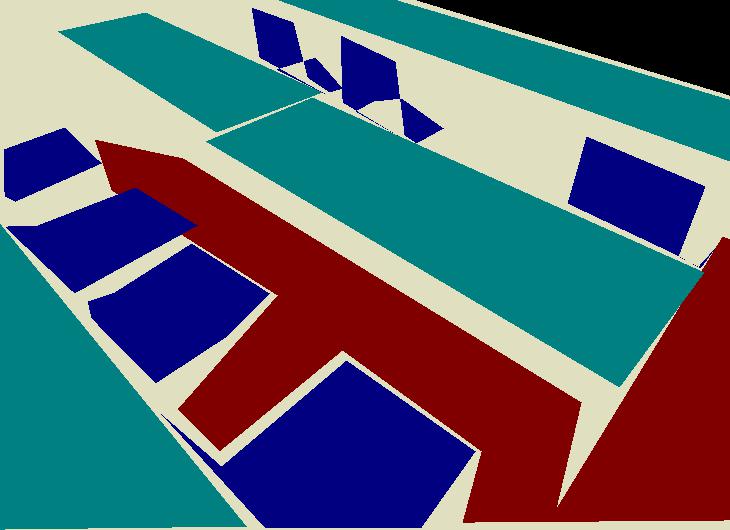}
		&\includegraphics[width=0.13\textwidth]{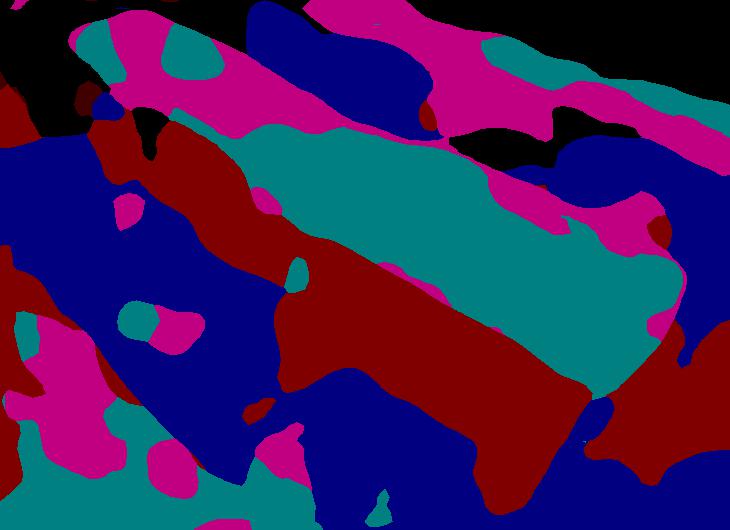}
		&\includegraphics[width=0.13\textwidth]{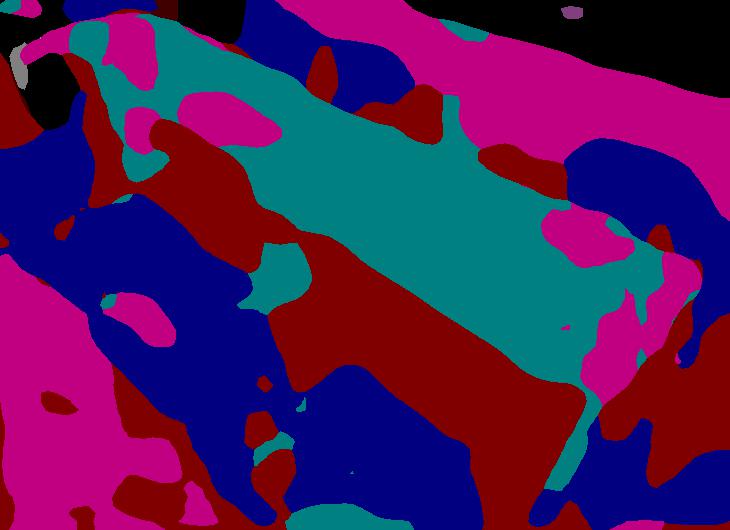}
		&\includegraphics[width=0.13\textwidth]{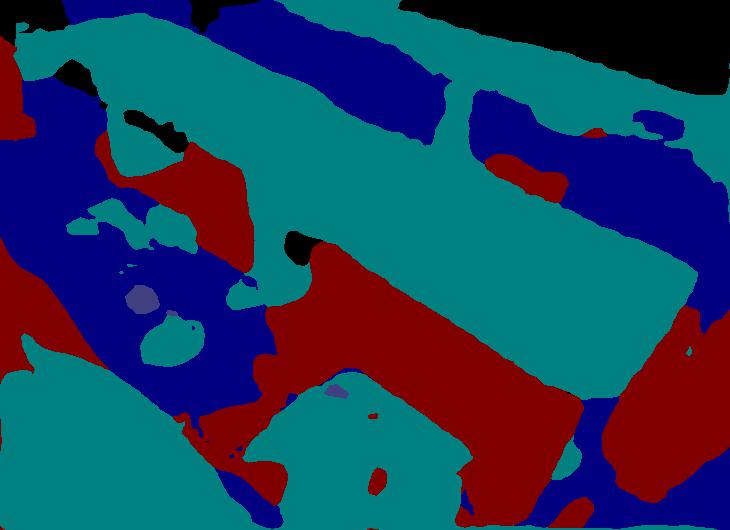}
		&\includegraphics[width=0.13\textwidth]{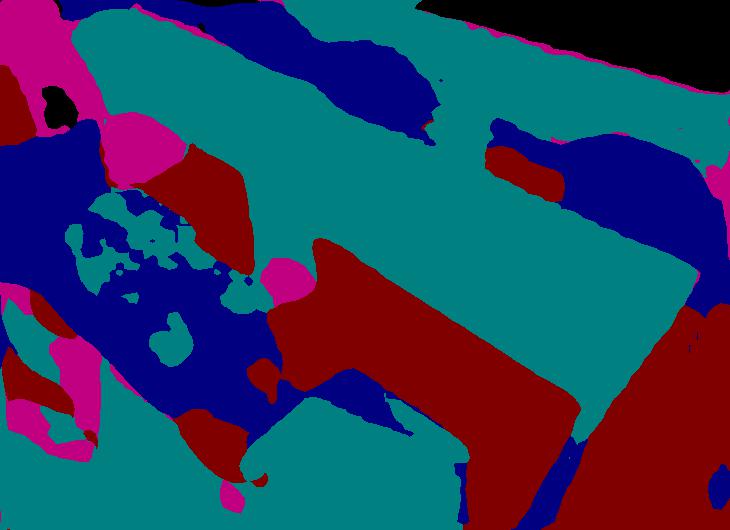}
		\\
		\includegraphics[width=0.13\textwidth]{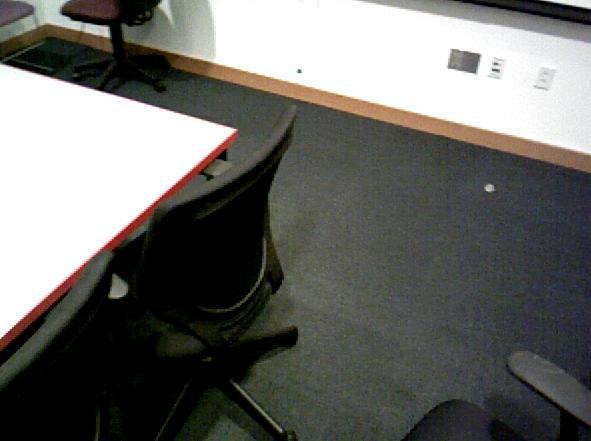}
		&\includegraphics[width=0.13\textwidth]{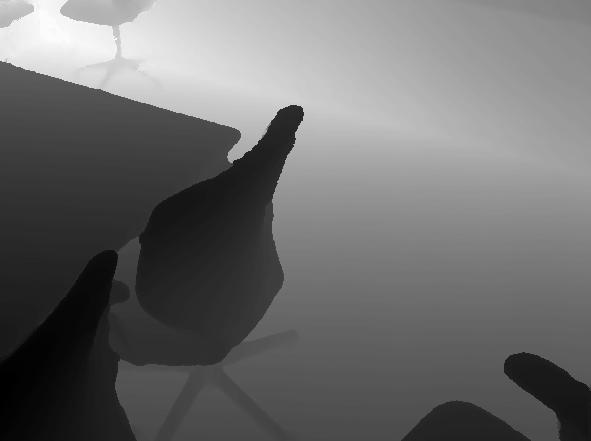}
		&\includegraphics[width=0.13\textwidth]{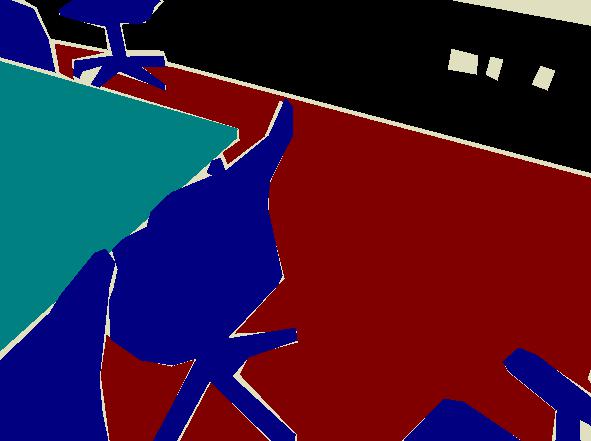}
		&\includegraphics[width=0.13\textwidth]{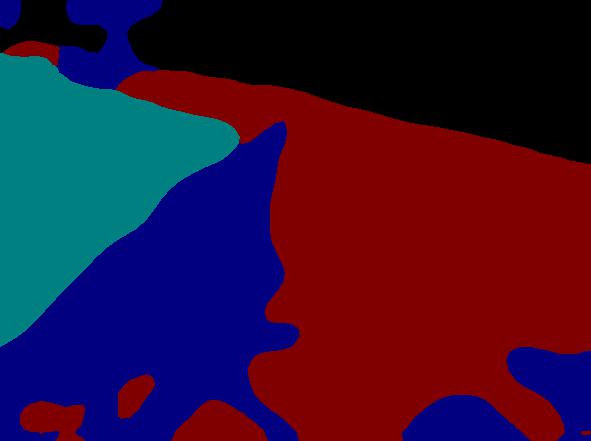}
		&\includegraphics[width=0.13\textwidth]{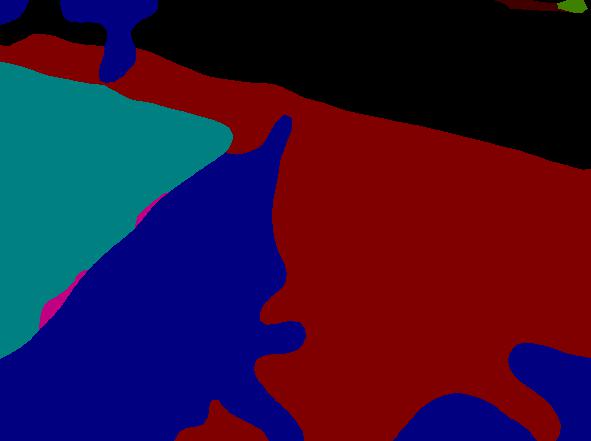}
		&\includegraphics[width=0.13\textwidth]{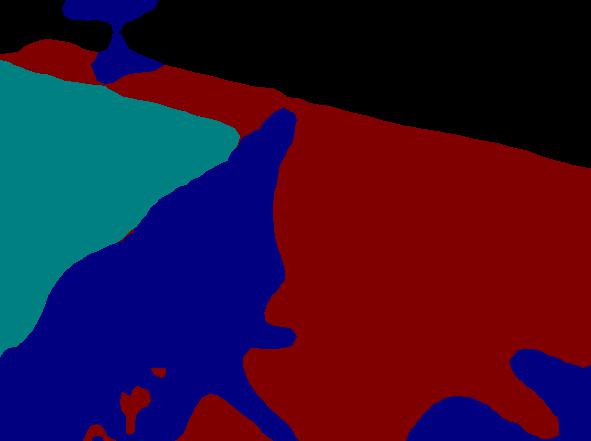}
		&\includegraphics[width=0.13\textwidth]{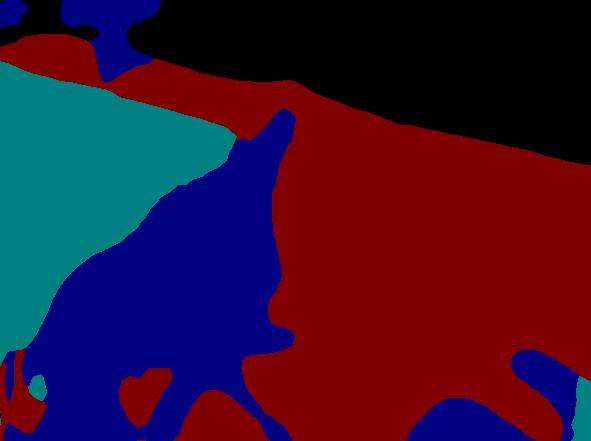}
		\\
		\includegraphics[width=0.13\textwidth]{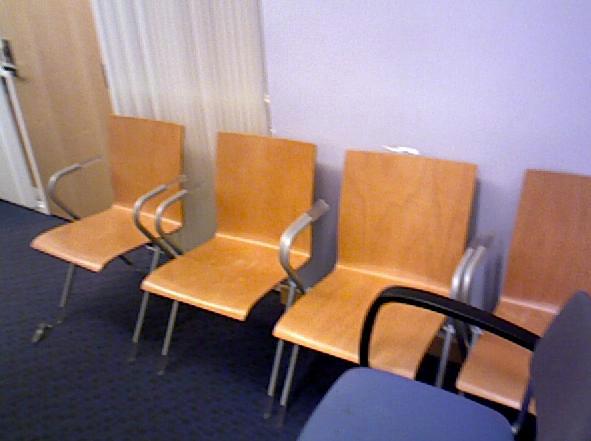}
		&\includegraphics[width=0.13\textwidth]{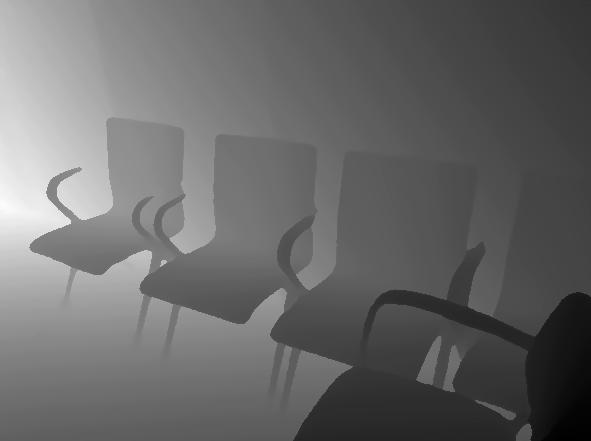}
		&\includegraphics[width=0.13\textwidth]{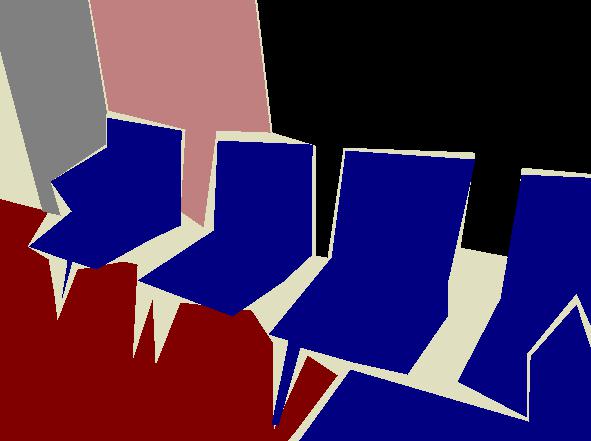}
		&\includegraphics[width=0.13\textwidth]{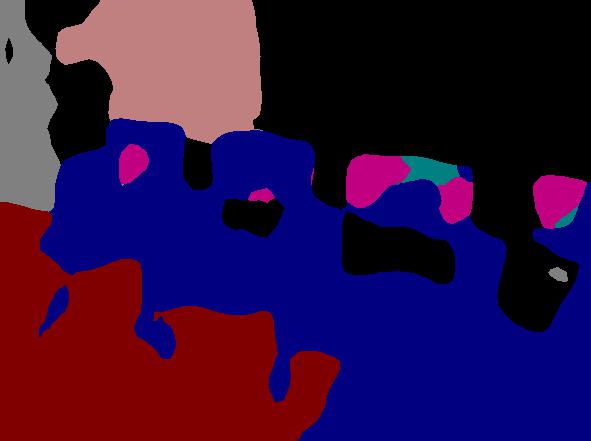}
		&\includegraphics[width=0.13\textwidth]{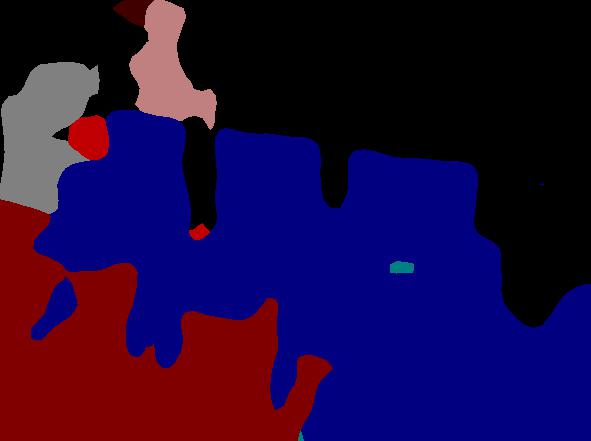}
		&\includegraphics[width=0.13\textwidth]{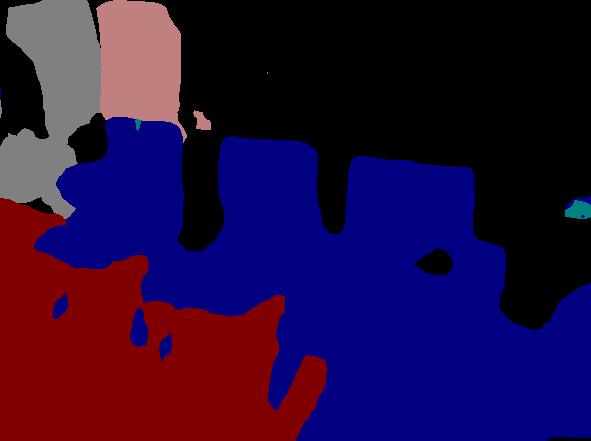}
		&\includegraphics[width=0.13\textwidth]{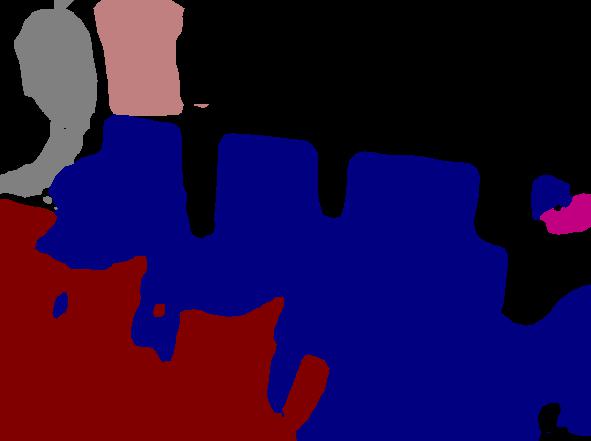}
		\\
		\includegraphics[width=0.13\textwidth]{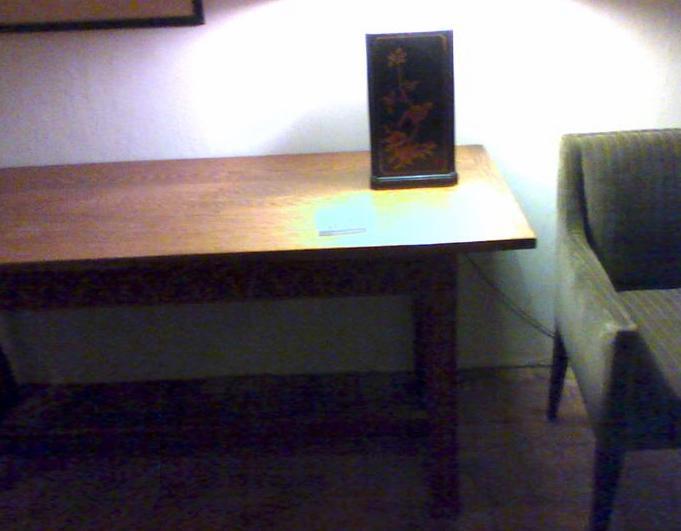}
		&\includegraphics[width=0.13\textwidth]{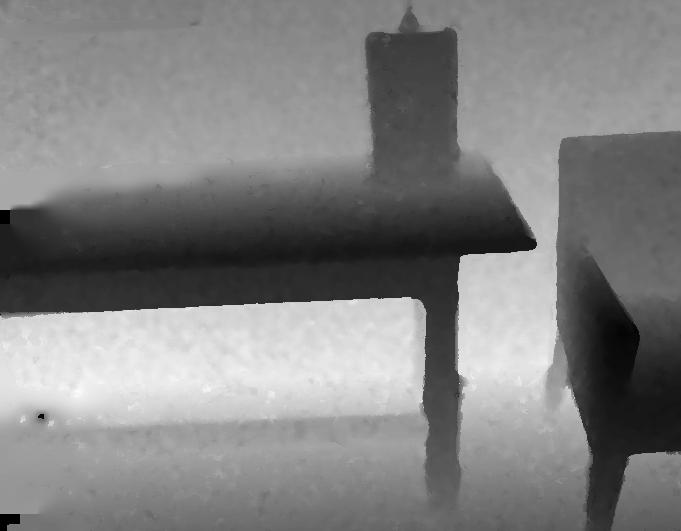}
		&\includegraphics[width=0.13\textwidth]{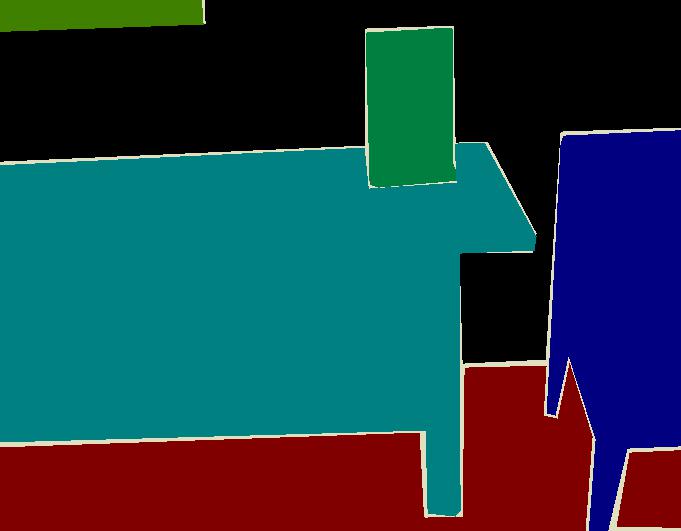}
		&\includegraphics[width=0.13\textwidth]{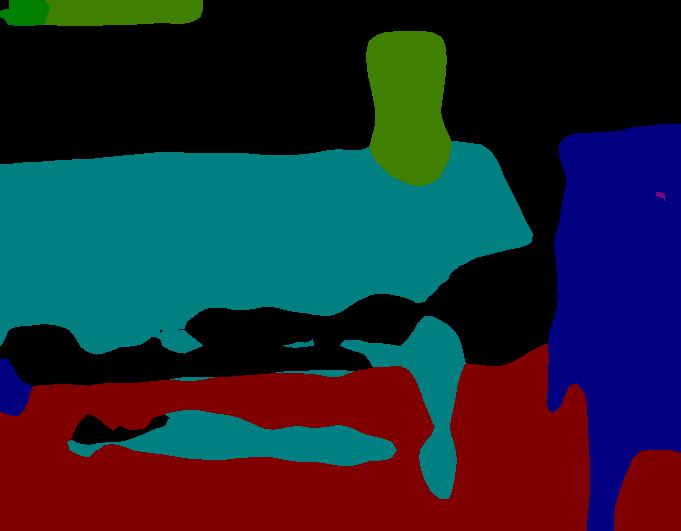}
		&\includegraphics[width=0.13\textwidth]{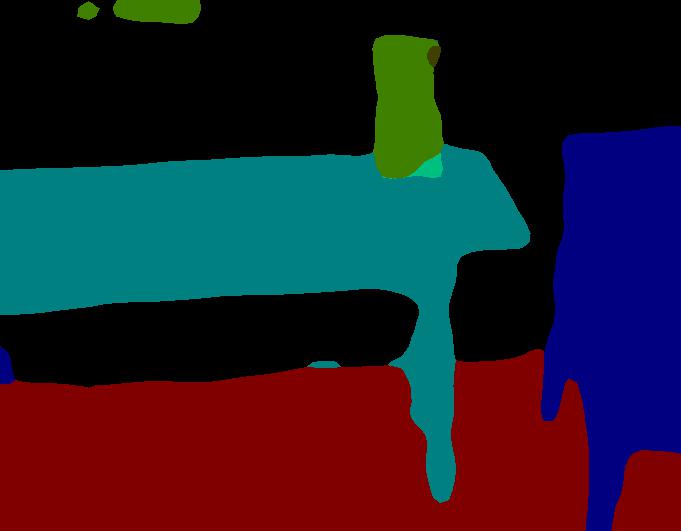}
		&\includegraphics[width=0.13\textwidth]{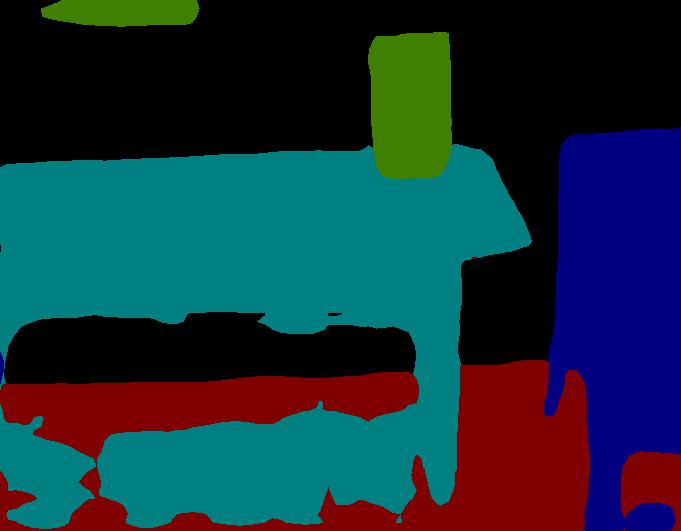}
		&\includegraphics[width=0.13\textwidth]{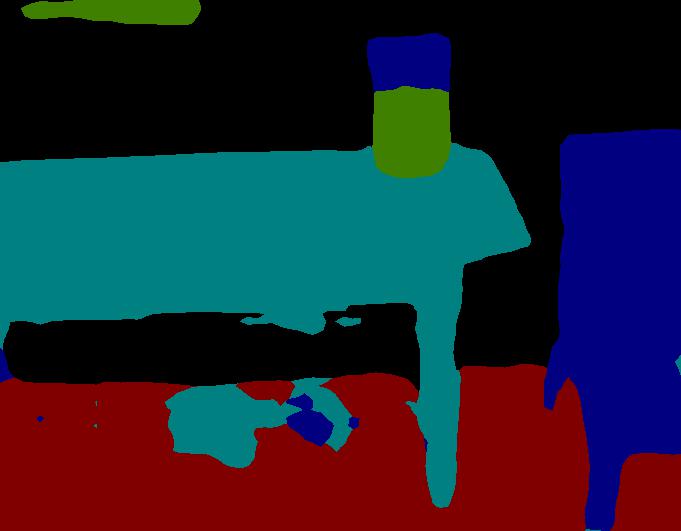}
		\\
		\includegraphics[width=0.13\textwidth]{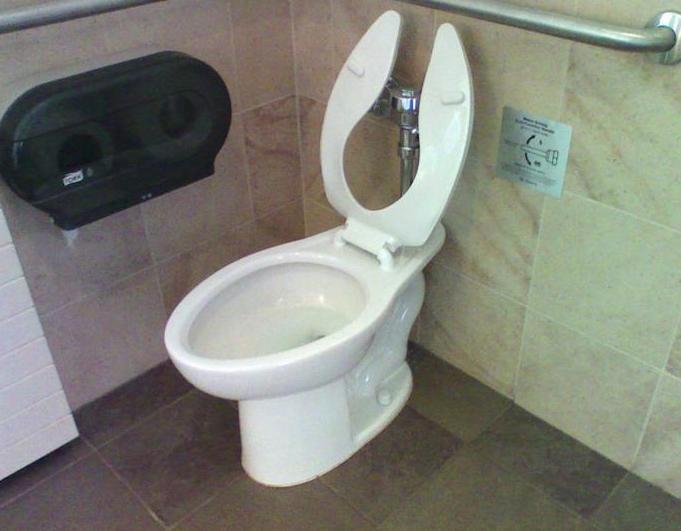}
		&\includegraphics[width=0.13\textwidth]{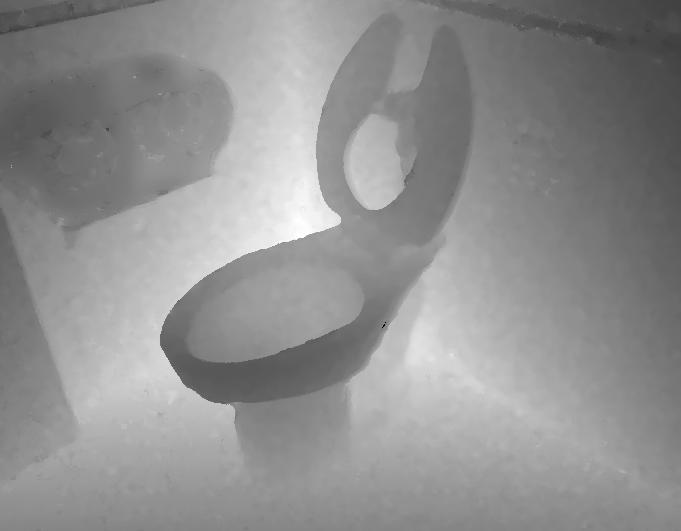}
		&\includegraphics[width=0.13\textwidth]{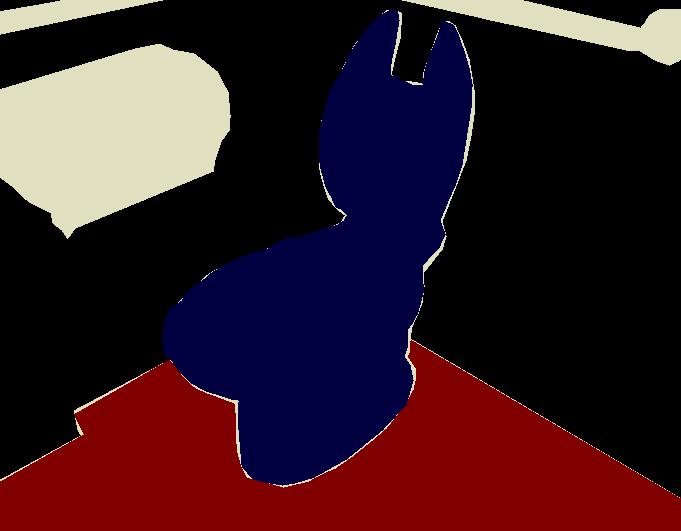}
		&\includegraphics[width=0.13\textwidth]{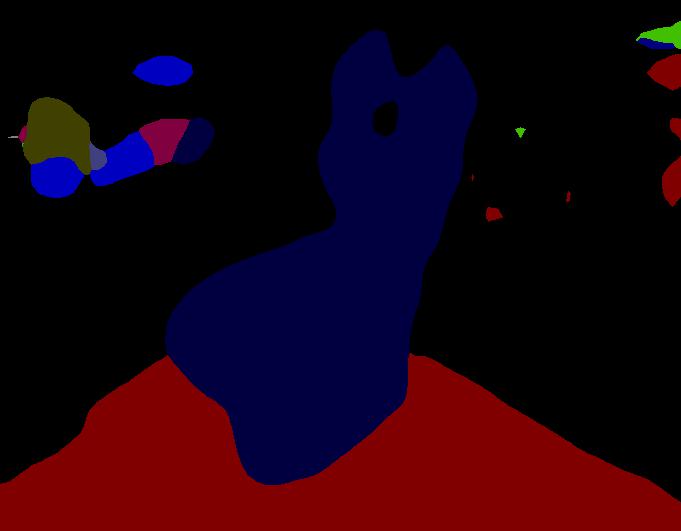}
		&\includegraphics[width=0.13\textwidth]{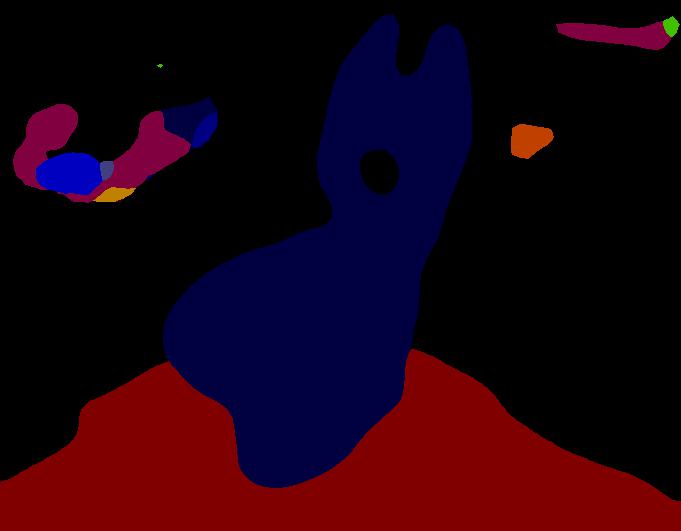}
		&\includegraphics[width=0.13\textwidth]{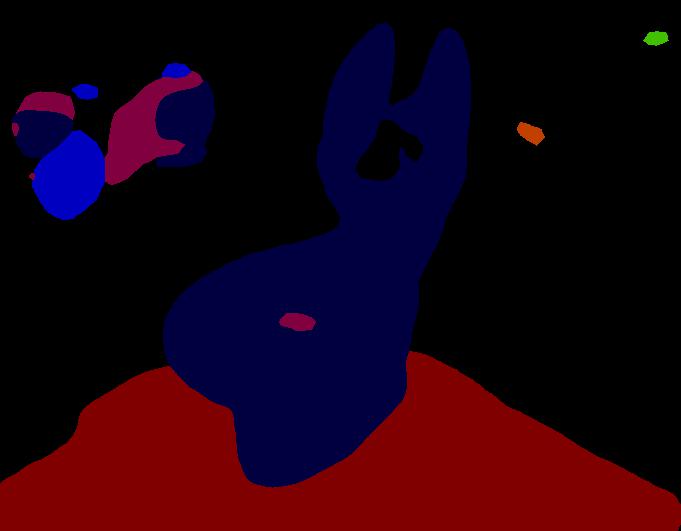}
		&\includegraphics[width=0.13\textwidth]{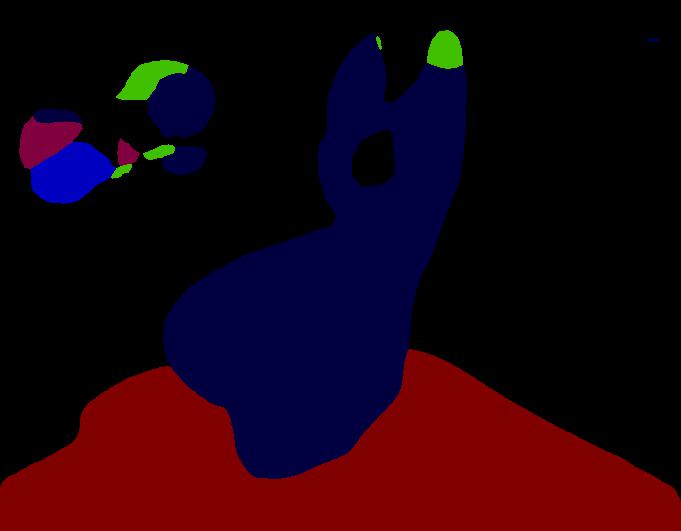}
		\\
		\tiny{RGB }& \tiny{Depth }& \tiny{GT }& \tiny{Baseline} & \tiny{HHA }&\tiny{D-CNN}& \tiny{DCNN+HHA}
	\end{tabular}
	\caption{Segmentation results on SUN-RGBD test dataset. ``GT" denotes ground truth. The white regions in ``GT" are the ignoring category. Networks are trained from pre-trained models.}
	\label{fig:sunrgbd}\vspace{-10pt}
\end{figure}
\paragraph{\bfseries{SID}} The comparison results on SID between D-CNN with its baseline are reported in Table~\ref{table:SID}. Networks are trained from scratch. Using depth images, D-CNN is able to achieve $4\%$ IoU over CNN while preserving the same number of parameters and computation complexity.
\vspace{-10pt}
\begin{table}
	\begin{center}
		\newcolumntype{C}{>{\centering\arraybackslash}p{3.5em}}
		\newcolumntype{E}{>{\centering\arraybackslash}p{7em}}
		\begin{tabular}{c|cc}
			\Xhline{3\arrayrulewidth}
			& Baseline&  D-CNN\\
			\hline
			Acc (\%)& 64.3&\bf{65.4}\\
			mAcc (\%)& 46.7&\bf{55.5}\\
			mIoU (\%)& 35.5&\bf{39.5}\\
			fwIoU (\%)& 48.5&\bf{49.9}\\
			\Xhline{3\arrayrulewidth}
		\end{tabular}
	\end{center}
	\caption{Comparison with baseline CNNs on SID Area $5$. Networks are trained from scratch.}\vspace{-40pt}
	\label{table:SID}
\end{table}

\subsection{Ablation Study}
\label{sec:ablation}
In this section, we conduct ablation studies on NYUv2 dataset to validate efficiency and efficacy of our approach. Testing results on NYUv2 test set are reported.

\paragraph{\bfseries{Depth-aware CNN}}
To verify the functionality of both depth-aware convolution and depth-aware average pooling, the following experiments are conducted.
\begin{itemize}
\item VGG-1: \texttt{Conv1\_1}, \texttt{Conv2\_1}, \texttt{Conv3\_1}, \texttt{Conv4\_1}, \texttt{Conv5\_1} and \texttt{Conv6} in VGG-16 are replaced with depth-aware convolution. This is the same as in Table~\ref{table:netarch}.
\item VGG-2: \texttt{Conv4\_1}, \texttt{Conv5\_1} and \texttt{Conv6} in VGG-16 are replaced with depth-aware convolution. Other layers remain the same as in Table~\ref{table:netarch}.
\item VGG-3: The depth-aware average pooling layer listed in Table~\ref{table:netarch} is replaced with regular pooling. Other layers remain the same as in Table~\ref{table:netarch}.
\end{itemize}
Results are shown in Table~\ref{table:depthconv}. Compared to VGG-2, VGG-1 adds depth-aware convolution in bottom layers. This helps the network propagate more fine-grained features with geometric relationships and increase segmentation performance by $6\%$ in IoU. The depth-aware average pooling operation is able to further promote the accuracy.   

\begin{table}
	\begin{center}
		\newcolumntype{C}{>{\centering\arraybackslash}p{3em}}
		\newcolumntype{E}{>{\centering\arraybackslash}p{5em}}
		\begin{tabular}{c|cc|ccc}
			\Xhline{3\arrayrulewidth}
			& Baseline& HHA& VGG-1    & VGG-2 & VGG-3  \\
			\hline
			Acc (\%)&50.1  &59.1&\bf{60.3} & 56.0  & 59.3  \\
			mAcc (\%)&23.9 &30.8&\bf{39.3} & 32.2  & 39.2    \\
			mIoU (\%)&15.9 &21.9&\bf{27.8} & 22.4  & 27.4     \\
			fwIoU (\%)&34.2&43.0&\bf{44.9} & 40.2  & 44.0     \\
			\Xhline{3\arrayrulewidth}
		\end{tabular}
	\end{center}
	\caption{Results of using depth-aware operations in different layers. Experiments are conducted on NYUv2 test set. Networks are trained from scratch.}
	\vspace{-10pt}
	\label{table:depthconv}
\end{table}

We also replace VGG-16 to ResNet-50 as the encoder. To build depth-aware ResNet, the \texttt{Conv3\_1}, \texttt{Conv4\_1}, and \texttt{Conv5\_1} in ResNet-50 are replaced with depth-aware convolution. The networks are initialized with parameters pre-trained on ADE20K~\cite{zhou2017scene}. Detailed architecture and training details for ResNet can be found in Supplementary Materials. Results are listed in Table~\ref{table:resnet}. 

\begin{table}
	\begin{center}
		\newcolumntype{C}{>{\centering\arraybackslash}p{3em}}
		\newcolumntype{E}{>{\centering\arraybackslash}p{5em}}
		\begin{tabular}{c|ccc}
			\Xhline{3\arrayrulewidth}
			& ResNet& ResNet-D-CNN& VGG-D-CNN  \\
			\hline
			Acc (\%)& 68.9 &\bf{69.6} &69.4 \\
			mAcc (\%)&50.2 &53.3 &\bf{53.6} \\
			mIoU (\%)&38.8 &\bf{41.5} &41.0 \\
			fwIoU (\%)&54.4 &54.4 &\bf{54.5} \\
			\Xhline{3\arrayrulewidth}
		\end{tabular}
	\end{center}
	\caption{Results of using depth-aware operations in ResNet-50. ``VGG-D-CNN" denotes the same network and result as in Table~\ref{table:nyud2}. Networks are trained from pre-trained models.}
		\vspace{-20pt}
	\label{table:resnet}
\end{table}

\paragraph{\bfseries{Depth Similarity Function}} We modify $\alpha$ and $F_\mathbf{D}$ to further validate the effect of different choices of depth similarity function on performance. We conduct the following experiments:

\begin{itemize}
	\item $\alpha_{8.3}$: $\alpha$ is set to $8.3$. The network architecture is the same as Table~\ref{table:netarch}. 
	\item $\alpha_{20}$: $\alpha$ is set to $20$. The network architecture is the same as Table~\ref{table:netarch}. 
	\item $\alpha_{2.5}$: $\alpha$ is set to $2.5$. The network architecture is the same as Table~\ref{table:netarch}. 
	\item clip$F_\mathbf{D}$: The network architecture is the same as Table~\ref{table:netarch}. $F_\mathbf{D}$ is defined as 
	\begin{equation}
	F_\mathbf{D}(\mathbf{p}_i,\mathbf{p}_j)=
	\begin{cases}
	0,& |\mathbf{D}(\mathbf{p}_i) - \mathbf{D}(\mathbf{p}_j)| \geq 1\\
	1,              & \text{otherwise}
	\end{cases}
	\label{eq:fd_clip}
	\end{equation}
\end{itemize}

Table~\ref{table:depthsim} reports the test performances with different depth similarity functions. Though the performance varies with different $\alpha$, they are all superior to baseline and even ``HHA". The result of clip$F_\mathbf{D}$ is also comparable with ``HHA". This validate that the effectiveness of using a depth-sensitive term to weight the contributions of neurons. \vspace{-10pt}
\begin{table}
\begin{center}
\newcolumntype{C}{>{\centering\arraybackslash}p{3em}}
\newcolumntype{E}{>{\centering\arraybackslash}p{5em}}
\begin{tabular}{c|cc|CCCC}
	\Xhline{3\arrayrulewidth}
 & Baseline& HHA&$\alpha_{8.3}$& $\alpha_{20}$ & $\alpha_{2.5}$ & clip$F_\mathbf{D}$\\
\hline
Acc (\%)&50.1  &59.1&\bf{60.3} &  58.5 &  58.5 &  53.0  \\
mAcc (\%)&23.9 &30.8&\bf{39.3} &  35.2 &  35.9 &  29.8   \\
mIoU (\%)&15.9 &21.9&\bf{27.8} &  24.9 &  25.3 &  20.1   \\
fwIoU (\%)&34.2&43.0&\bf{44.9} &  42.6 &  42.9 &  37.5   \\
\Xhline{3\arrayrulewidth}
\end{tabular}
\end{center}
\caption{Results of using different $\alpha$ and $F_\mathbf{D}$. Experiments are conducted on NYUv2 test set. Networks are trained from scratch.}
	\vspace{-40pt}
\label{table:depthsim}
\end{table}

\begin{figure}
	\centering
	\begin{tabular}{cc}
	\includegraphics[width=.65\textwidth]{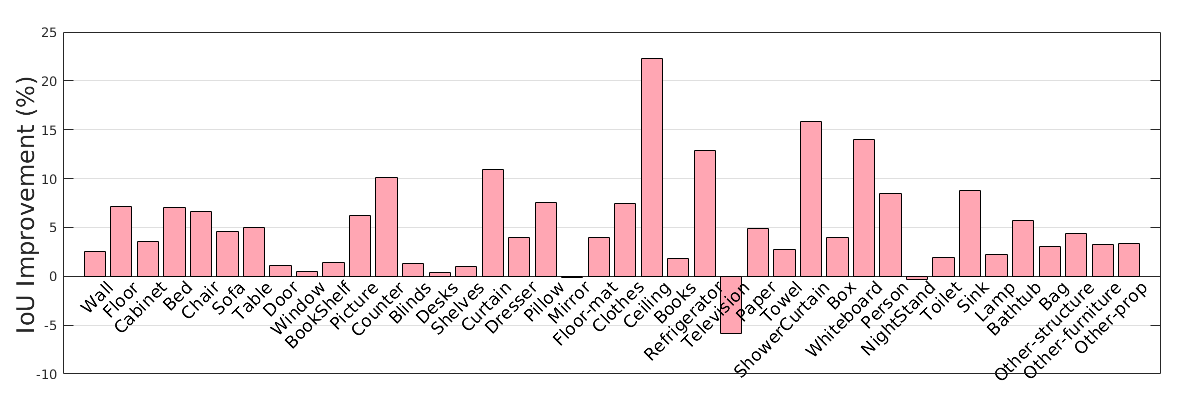}
	&
	\includegraphics[width=.3\textwidth]{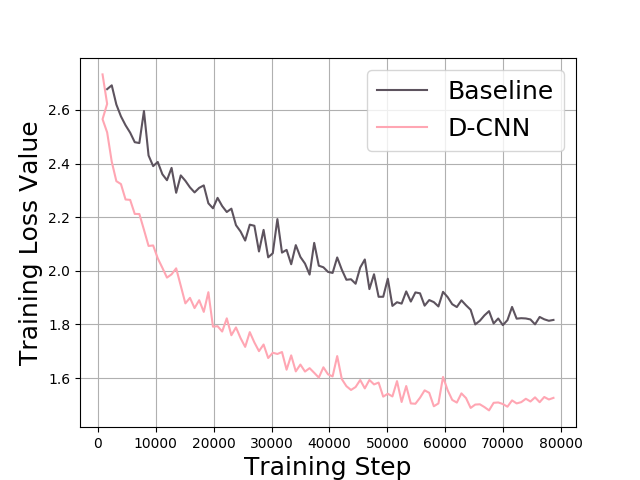}\\
%	Average per-class IoU improvement of D-CNN over baseline.&Training loss\\
	(a)&(b)
	\end{tabular}
	\vspace{-10pt}
	\caption{Performance Analysis. (a) Per-class IoU improvement of D-CNN over baseline on NYUv2 test dataset. (b) Evolution of training loss on NYUv2 train dataset. Networks are trained from scratch.}
	\vspace{-15pt}
	\label{fig:breakdown}
\end{figure}

\paragraph{\bfseries{Performance Analysis}} To have a better understanding of how depth-aware CNN outperforms the baseline, we visualize the improvement of IoU for each semantic class in Figure~\ref{fig:breakdown}(a). The statics shows that D-CNN outperform baseline on most object categories, especially these large objects such as ceilings and curtain. Moreover, we observe depth-aware CNN has a faster convergence than baseline, especially trained from scratch. Figure~\ref{fig:breakdown}(b) shows the training loss evolution with respect to training steps. Our network gains lower loss values than baseline.

%\begin{figure}
%	\centering
%	\includegraphics[width=\textwidth]{figures/trainingloss.png}
%	\caption{Average per-class IoU improvement of D-CNN over baseline.}
%	\label{fig:trainingloss}
%\end{figure}

\subsection{Model Complexity and Runtime Analysis}
\label{sec:time}
Table~\ref{table:time} reports the model complexity and runtime of D-CNN and the state-of-the-art method~\cite{xiaojuaniccv17}. In their method, kNN takes $O(kN)$ runtime at least, where N is the number of pixels. We leverage the grid structure of raw depth input. Without increasing any model parameters, D-CNN is able to incorporate geometric information in CNN efficiently.
\vspace{-10pt}

\begin{table}
\begin{center}
\newcolumntype{C}{>{\centering\arraybackslash}p{3.5em}}
\newcolumntype{E}{>{\centering\arraybackslash}p{5em}}
\begin{tabular}{c|EC|CCCC}
	\Xhline{3\arrayrulewidth}
& Baseline & HHA & \cite{xiaojuaniccv17} & D-CNN \\
\hline
net. forward (ms)& 32.5 & 64.2 & 214 & 39.3\\
\# of params & 47.0M & 92.0M & 47.25M & 47.0M\\
\Xhline{3\arrayrulewidth}
\end{tabular}
\end{center}
\caption{Model complexity and runtime comparison. Runtime is tested on Nvidia 1080Ti, with input image size $425\times 560\times 3$. }
	\vspace{-40pt}
\label{table:time}
\end{table}

\section{Conclusion}
\vspace{-5pt}
We present a novel depth-aware CNN by introducing two operations: depth-aware convolution and depth-aware average pooling. Depth-aware CNN augments conventional CNN with a depth similarity term and encode geometric variance into basic convolution and pooling operations. By adapting effective receptive field, these depth-aware operations are able to incorporate geometry into CNN while preserving CNN's efficiency. Without introducing any parameters and computational complexity, this method is able to improve the performance on RGB-D segmentation over baseline by a large margin. Moreover, depth-aware CNN is flexible and easily replaces its plain counterpart in standard CNNs. Comparison with the state-of-the-art methods and extensive ablation studies on RGB-D semantic segmentation demonstrate the effectiveness and efficiency of depth-aware CNN.

Depth-aware CNN provides a general framework for vision tasks with RGB-D input. 
Moreover, depth-aware CNN takes the raw depth image as input and bridges the gap between 2D CNN and 3D geometry. 
In future works, we will apply depth-aware CNN on various tasks such as 3D detection, instance segmentation and we will perform depth-aware CNN on more challenging dataset. 
Apart from depth input, we will exploit various 3D data such as LiDAR point cloud.

\section*{Acknowledgements}
We thank Ronald Yu, Yi Zhou and Qiangui Huang for the discussion and proofread.
This research is supported by the Intelligence Advanced Research Projects Activity (IARPA) via
Department of Interior/ Interior Business Center (DOI/IBC) contract number
D17PC00288. The U.S. Government is authorized to reproduce and distribute reprints
for Governmental purposes notwithstanding any copyright annotation thereon.
Disclaimer: The views and conclusions contained herein are those of the authors and
should not be interpreted as necessarily representing the official policies or
endorsements, either expressed or implied, of IARPA, DOI/IBC, or the U.S.
Government.

\bibliographystyle{splncs}
\bibliography{egbib}
\end{document}